\documentclass[lettersize,journal]{IEEEtran}
\usepackage{amsmath,amsfonts}
\usepackage{algorithmic}
\usepackage{algorithm}
\usepackage{array}
\usepackage{textcomp}
\usepackage{stfloats}
\usepackage{url}
\usepackage{verbatim}
\usepackage{graphicx}
\usepackage{cite}
\usepackage{amsmath}
\usepackage{amssymb}
\usepackage[table]{xcolor}
\usepackage[pagebackref,breaklinks,colorlinks]{hyperref}
\usepackage[group-separator={,}]{siunitx}
\usepackage{multirow}
\usepackage{tabularx}
\usepackage{booktabs} 
\usepackage{makecell}
\usepackage{graphbox}
\usepackage{footmisc}
\usepackage{bm}
\usepackage{caption}
\usepackage{arydshln}
\usepackage{dashrule}
\usepackage{subcaption}
\usepackage{enumitem}
\setlist[itemize]{noitemsep,nolistsep}

\usepackage[capitalize]{cleveref}
\crefname{section}{Sec.}{Secs.}
\Crefname{section}{Section}{Sections}
\Crefname{table}{Table}{Tables}
\crefname{table}{Tab.}{Tabs.}
\Crefname{figure}{Figure}{Figures}
\crefname{figure}{Fig.}{Figs.}
\Crefname{equation}{Equation}{Equations}
\crefname{equation}{Eq.}{Eqs.}

\usepackage{xspace}
\makeatletter
\DeclareRobustCommand\onedot{\futurelet\@let@token\@onedot}
\def\@onedot{\ifx\@let@token.\else.\null\fi\xspace}

\def\ie{\emph{i.e}\onedot} 
 
\def\etc{\emph{etc}\onedot} 
 
\def\etal{\emph{et al}\onedot}
\makeatother

\usepackage{dsfont}

\newcommand{\R}{\mathbb{R}}
\newcommand{\F}{\mathbf{F}}
\newcommand{\G}{\mathbf{G}}
\newcommand{\Q}{\mathbf{Q}}
\renewcommand{\S}{\mathbf{S}}
\newcommand{\K}{\mathbf{K}}
\newcommand{\V}{\mathbf{V}}
\newcommand{\1}{\mathds{1}}

\colorlet{lightpink}{pink!35}
\colorlet{lightcyan}{cyan!20}
\colorlet{red}{red!80}
\colorlet{blue}{blue!80}
\colorlet{green}{green!60!black}

\usepackage{pifont}
\newcommand{\cmark}{\ding{51}}%

\newcommand{\best}[1]{\textcolor{red}{\textbf{#1}}}
\newcommand{\second}[1]{\textcolor{blue}{\textbf{#1}}}

\newcolumntype{C}[1]{>{\centering\arraybackslash}p{#1}}
\newcolumntype{L}[1]{>{\raggedleft\arraybackslash}p{#1}}
\newcolumntype{R}[1]{>{\raggedright\arraybackslash}p{#1}}

\newcommand{\RNum}[1]{\uppercase\expandafter{\romannumeral #1\relax}}

\newcommand{\circled}[1]{\textcircled{\raisebox{-0.8pt}{#1}}}

\newcommand{\ours}{TOPIQ\xspace}

\graphicspath{{images/}}

\usepackage[switch]{lineno}
\begin{document}

\title{TOPIQ: A Top-down Approach from Semantics to Distortions for Image Quality Assessment}

\author{Chaofeng Chen,
        Jiadi Mo,
        Jingwen Hou,~\IEEEmembership{Student Member,~IEEE}, 
        Haoning Wu,
        Liang Liao,~\IEEEmembership{Member,~IEEE},
        Wenxiu Sun,
        Qiong Yan,
        Weisi Lin,~\IEEEmembership{Fellow,~IEEE}
\thanks{C. Chen, J. Mo, J. Hou, H. Wu, L. Liao and W. Lin are with School of Computer Science and Engineering, Nanyang Technological University, Singapore. (Email: [chaofeng.chen, liang.liao, wslin]@ntu.edu.sg, [JMO004, jingwen003, haoning001]@e.ntu.edu.sg)\protect\\
\indent\indent W. Sun and Q. Yan are with Tetras. AI and Sensetime Research. (Email: [irene.wenxiu.sun,  sophie.yanqiong]@gmail.com)\protect\\
\indent\indent Corresponding author: Weisi Lin.}}

\markboth{Journal of \LaTeX\ Class Files,~Vol.~14, No.~8, August~2021}%
{Shell \MakeLowercase{\textit{et al.}}: A Sample Article Using IEEEtran.cls for IEEE Journals}


\maketitle

\begin{abstract}
Image Quality Assessment (IQA) is a fundamental task in computer vision that has witnessed remarkable progress with deep neural networks. Inspired by the characteristics of the human visual system, existing methods typically use a combination of global and local representations (\ie, multi-scale features) to achieve superior performance. However, most of them adopt simple linear fusion of multi-scale features, and neglect their possibly complex relationship and interaction. In contrast, humans typically first form a global impression to locate important regions and then focus on local details in those regions. We therefore propose a top-down approach that uses high-level semantics to guide the IQA network to focus on semantically important local distortion regions, named as \emph{TOPIQ}. Our approach to IQA involves the design of a heuristic coarse-to-fine network (CFANet) that leverages multi-scale features and progressively propagates multi-level semantic information to low-level representations in a top-down manner. A key component of our approach is the proposed cross-scale attention mechanism, which calculates attention maps for lower level features guided by higher level features. This mechanism emphasizes active semantic regions for low-level distortions, thereby improving performance. CFANet can be used for both Full-Reference (FR) and No-Reference (NR) IQA. We use ResNet50 as its backbone and demonstrate that CFANet achieves better or competitive performance on most public FR and NR benchmarks compared with state-of-the-art methods based on vision transformers, while being much more efficient (with only ${\sim}13\%$ FLOPS of the current best FR method). Codes are released at \url{https://github.com/chaofengc/IQA-PyTorch}.

\end{abstract}


\section{Introduction}

\begin{figure}[t]
    \centering
    \includegraphics[width=0.49\linewidth]{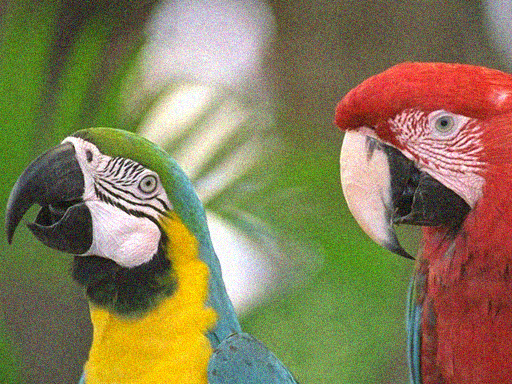}
    \includegraphics[width=0.49\linewidth]{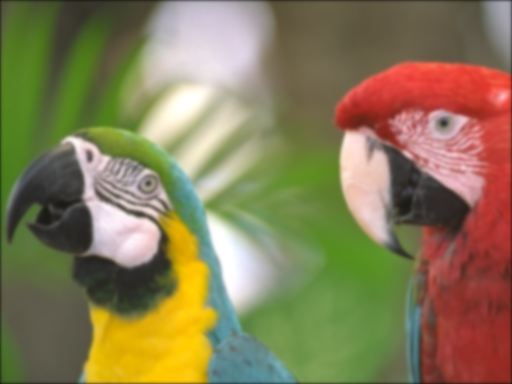} 
    \\
    \makebox[0.49\linewidth]{Distorted image A}
    \makebox[0.49\linewidth]{Distorted image B}
    \\[0.5em]
    \begin{tabularx}{\linewidth}{X|C{1.8cm}C{1.8cm}}
     &  A is better & B is better \\ \hline
     Humans (MOS) &  \cmark & \\
     PSNR, SSIM, MS-SSIM &  & \cmark \\
     LPIPS, DISTS & & \cmark \\ 
     Ours & \textcolor{red}{\cmark} & 
    \end{tabularx}
    \caption{An example from the TID2013 dataset \cite{ponomarenko2013tid2013} (the reference image is omitted for easier comparison). It is noticeable that, although the large background region is noisy in image A, humans assign a higher quality score (Mean Opinion Score, a.k.a., MOS) to A than to B, because the birds' region in A is much clearer. This indicates that humans tend to focus on more semantically important regions. Simple multi-scale approaches such as LPIPS and DISTS ignore the correlation between high-level semantics and low-level distortions, and therefore, produce inconsistent judgments compared to humans.}
    \label{fig:intro_motivation}
\end{figure}

\IEEEPARstart{I}{mage} Quality Assessment (IQA) aims to estimate perceptual image quality similar to the human visual system (HVS). It can be useful in enhancing the visual experience of humans in various applications such as image acquisition, compression, restoration, editing, and generation. The rapid advancement of image processing algorithms based on deep learning has created an urgent need for better IQA metrics.

\begin{figure*}[t]
    \centering
    \begin{subfigure}[t]{0.29\linewidth}
        \centering
        \includegraphics[width=\linewidth]{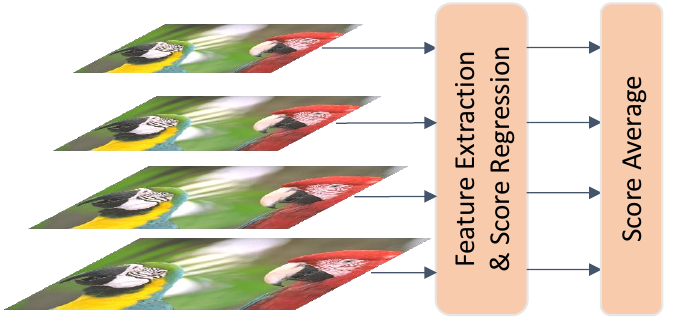}
        \caption{Image pyramid: parallel approach, such as MS-SSIM, NIQE \etc.} \label{fig:arch_parrallel}
    \end{subfigure}
    \hfill
    \begin{subfigure}[t]{0.29\linewidth}
        \centering
        \includegraphics[width=\linewidth]{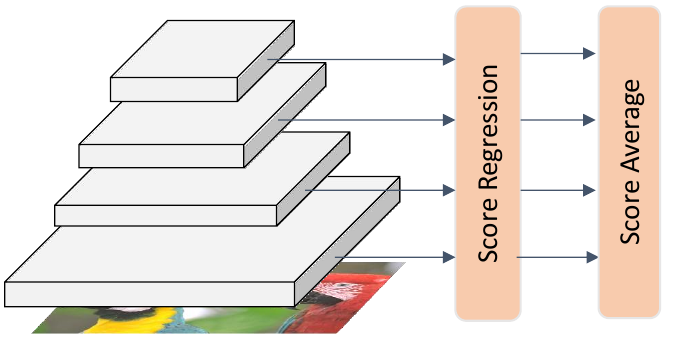}
        \caption{Feature pyramid: bottom-up approach, such as LPIPS, DISTS \etc.} \label{fig:arch_btup}
    \end{subfigure}
    \hfill
    \begin{subfigure}[t]{0.40\linewidth}
        \centering
        \includegraphics[width=\linewidth]{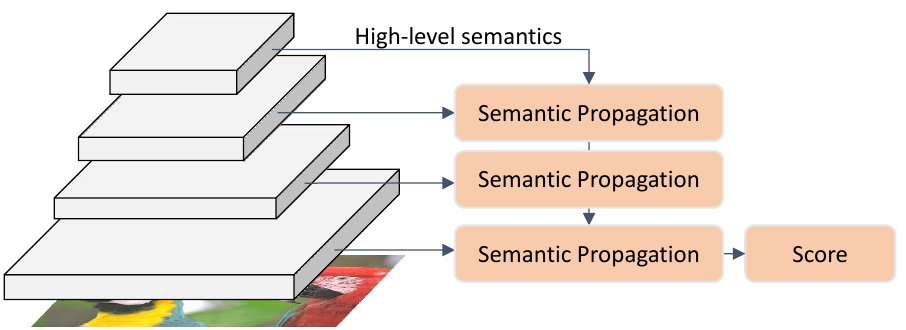}
        \caption{Feature pyramid: \textbf{our top-down approach}.} \label{fig:arch_topdown}
    \end{subfigure}
    \caption{Three types of IQA framework based on how they extract and employ multi-scale features: the parallel, bottom-up and top-down methods.}
    \label{fig:framework-classify}
\end{figure*}

According to the requirement for pristine reference images, most IQA techniques can be categorized as Full-Reference (FR) IQA or No-Reference (NR) IQA. In both cases, multi-scale feature extraction is a crucial method to enhance the performance and is commonly utilized in both hand-crafted and deep learning features. These multi-scale techniques can be roughly classified into three categories based on how they extract and use multi-scale features: the parallel, bottom-up, and top-down methods (as depicted in \cref{fig:framework-classify} for a brief overview).

Traditional approaches, such as MS-SSIM~\cite{msssim2003} and NIQE~\cite{2012niqe}, typically use the parallel paradigm (\cref{fig:arch_parrallel}). They resize the original image to create multi-scale inputs, and then extract features and calculate quality scores in parallel on these resized images. However, directly extracting features from multi-scale RGB images is often less effective because it is difficult to obtain meaningful quality representations from a low-resolution RGB image. Bottom-up approaches extract feature pyramids from original images in a bottom-up manner, such as the traditional steerable pyramid used in CW-SSIM \cite{cwssim2009}. Deep learning-based approaches, such as LPIPS \cite{zhang2018lpips} and DISTS \cite{dists2020}, naturally follow the bottom-up approach (\cref{fig:arch_btup}). They use features from different levels as individual components and estimate quality scores for them separately, and the final scores are obtained through a weighted sum. Although bottom-up approaches are more effective than parallel methods in extracting multi-scale features, they have similar drawbacks: 1) they do not consider the fact that high-level semantic information can guide the network to focus on more semantically active low-level features; 2) two images with different distortions may have similar high-level semantic features, making it difficult to use these features to regress quality scores directly. For example, in \cref{fig:intro_motivation}, image A has clearer bird heads but a much noisier background than image B. Humans are more sensitive to the quality of bird regions and tend to prefer image A, while MS-SSIM, LPIPS, and DISTS give better quality scores to image B due to the distraction from the large background region. This observation suggests that a top-down approach to exploiting multi-scale features, where high-level semantic features guide the level of distortion perception, may be beneficial (see \cref{fig:arch_topdown} for an example). However, to the best of our knowledge, most CNN-based approaches, including the latest works in the NTIRE IQA challenge \cite{gu2022ntire}, still follow the bottom-up paradigm, and the top-down approach for multi-scale features remains largely under-explored.

In this paper, we propose a top-down approach for IQA that utilizes deep multi-scale features. Our approach involves a heuristic coarse-to-fine attention network, referred to as CFANet. It emulates the process of the human visual system (HVS) by propagating semantic information from the highest level to the lowest level in a progressive manner. This heuristic design avoids the complexity of selecting among multiple features from different scales and has proven to be effective. Our key innovation is a novel cross-scale attention (CSA) mechanism that allows information propagation between different levels. The CSA takes high-level features as guidance to select important low-level distortion features. Inspired by the widely used attention mechanism in transformers \cite{vaswani2017attention}, the proposed CSA is formulated as a query problem based on feature similarities where high-level features serve as \emph{queries} and low-level features make \emph{(key, value)} pairs. Intuitively, the high level semantic features can be regarded as clustering centers, thereby aggregating low-level features that are more semantically active. We apply multiple CSA blocks to multi-scale features from pretrained CNN backbones, such as ResNet50 \cite{resnet2016}.

A practical challenge is that the spatial size of feature maps, increases quadratically from coarse to fine level, which makes it expensive to directly calculate cross-scale attention in the original multi-scale features. To address this, we introduce a gated local pooling (GLP) block to reduce the size of low-level features. The GLP block consists of a gated convolution followed by average pooling with a predefined window size. It helps filter out redundant information and significantly reduces the computational cost. We conduct comprehensive experimental comparisons on both FR and NR (including aesthetic) IQA datasets. Our CFANet demonstrates better or competitive performance with lower computational complexity.

Our contributions can be summarized as follows:
\begin{itemize}
    \item We introduce a top-down approach that leverages deep multi-scale features for IQA. Unlike previous parallel and bottom-up methods, our proposed CFANet can effectively propagate high-level semantic information from coarse to fine scales, enabling the network to focus on distortion regions that are more semantically important.
    \item We propose a novel cross-scale attention (CSA) mechanism to transfer high-level semantics to low-level distortion representations. Additionally, we introduce a gated local pooling (GLP) block that reduces the computational cost by filtering redundant information.
    \item Our proposed CFANet is significantly more efficient than state-of-the-art approaches. With a simple ResNet50 \cite{resnet2016} backbone, it achieves competitive performance while only requiring approximately 13\% of the floating point operations (FLOPS) of the best existing FR method.
\end{itemize}

\section{Related Works}

\subsection{Full-Reference Image Quality Assessment} 

FR-IQA methods compare a reference image and a distorted image to measure the dissimilarities between them. The most commonly used traditional metric is peak signal-to-noise ratio (PSNR), which is simple to calculate and represents the pixel-wise fidelity of the images. However, the HVS is highly non-linear, and the pixel-wise comparison of PSNR does not align with human perception. To address this, Wang \etal~\cite{ssim2004} introduced the structural similarity (SSIM) index to compare structural similarity in local patches, which inspired a lot of follow-up works \cite{sheikh2006vif,cwssim2009,larson2010mad,zhang2011fsim,xue2013gmsd,zhang2014vsi,laparra2016nlpd}. These works introduce more complicated hand-crafted features to measure image dissimilarities. 

Learning-based approaches have been proposed recently to overcome the limitations of hand-crafted features. However, early end-to-end works \cite{kim2017deepqa,bosse2017wadiqam} suffer from over-fitting. Zhang \etal \cite{zhang2018lpips} proposed a large-scale dataset and found that pretrained deep features are effective for measuring perceptual similarity. Similarly, Prashnani \etal \cite{Prashnani_2018_PieAPP} created a comparable dataset. Gu \etal \cite{pipal} proposed the PIPAL dataset and initiated the NTIRE2021~\cite{gu2021ntire} and NTIRE2022~\cite{gu2022ntire} IQA challenges. This greatly advanced deep learning-based IQA, leading to the emergence of many new approaches. Among these, methods based on vision transformers, such as IQT~\cite{cheon2021iqt} and AHIQ~\cite{lao2022ahiq}, perform the best.

\subsection{No-Reference Image Quality Assessment}
NR-IQA is a more challenging task due to a lack of reference images. There are two subtasks in NR-IQA: technical quality assessment~\cite{koniq10k} and aesthetic quality assessment~\cite{murray2012ava}. The former focuses on technical aspects of the image such as sharpness, brightness, and noise, and is commonly used to measure the fidelity of an image to the original scene and the accuracy of image acquisition, transmission, and reproduction. The latter, on the other hand, is concerned with the subjective perceptions of viewers towards the visual appeal of an image, taking into account aesthetic aspects such as composition, lighting, color harmony, and overall artistic impression. As such, image aesthetic evaluation is more subjective than image quality evaluation, as it is largely dependent on individual viewer's personal preferences and cultural background. Although they have different focus, both of them involve subjective or objective assessment of visual images, and are influenced by factors such as lighting, color accuracy, and sharpness. Traditional approaches for NR-IQA rely on natural scene statistics (NSS) \cite{moorthy2011nssdiivine,2012niqe,2012brisque,zhang2015ilniqe,ma2017nrqm,blau2018pi}. While NSS-based methods perform well in distinguishing synthetic technical distortions, they struggle with modeling authentic technical distortions and aesthetic quality assessment. As a result, many works have turned to deep learning for NR-IQA. They are generally improved with more advanced network architecture, from deep belief net~\cite{hou2014blind} to CNN~\cite{kang2014cnniqa}, then to deeper CNN~\cite{ma2017meon,kwanyee2018hiqa,2020dbcnn}, later to ResNet~\cite{zeng2018pqr,zhu2020metaiqa,hyperiqa}, and now vision transformers~\cite{tiqa2021,tres2022wacv,yang2022maniqa}. In additional to these works, there have been several notable works in NR-IQA. Liu \etal \cite{liu2017rankiqa} introduced a ranking loss for pretraining networks with synthetic data. Talebi \etal \cite{talebi2018nima} proposed a new distribution loss to replace simple score regression. Zheng \etal \cite{zheng2021ckdn} proposed generating the degraded-reference representation from the distorted image via knowledge distillation. Ke \etal \cite{ke2021musiq} employed multi-scale inputs and a vision transformer backbone to process images with varying sizes and aspect ratios. Hu \etal~\cite{hu2020subjective} focus on the quality evaluation of image restoration algorithms. They proposed a pairwise-comparison-based rank learning framework~\cite{hu2019pairwise} and a hierarchical discrepancy learning model~\cite{hu2022hierarchical} for performance benchmarking of image restoration algorithms.

Despite achieving promising performance, the latest approaches based on transformers are typically more computationally expensive than ResNet models to achieve the same level of performance with the same input size. Furthermore, the computational cost of transformers increases quadratically with larger image sizes, which can be a significant drawback. This work shows that by imitating the global-to-local process of the HVS, our model can achieve better or comparable performance in both FR and NR tasks using a simple ResNet50 as the backbone.

\section{The Top-Down Approach for IQA}

\subsection{Observations and Motivation} \label{sec:motivation}

To illustrate our motivation, we conducted a detailed analysis of two seminal multi-scale approaches: the MS-SSIM and LPIPS\footnote{LPIPS has many different versions. We use the VGG backbone of the latest 0.1 version here.}. We used example images from \cref{fig:intro_motivation} and the TID2013 dataset for our analysis.

\begin{figure}[t]
    \begin{subfigure}[t]{.99\linewidth}
        \raisebox{0.1\height}{\makebox[0.01\textwidth]{\rotatebox{90}{\makecell{\small Image A}}}}
        \includegraphics[width=.99\textwidth]{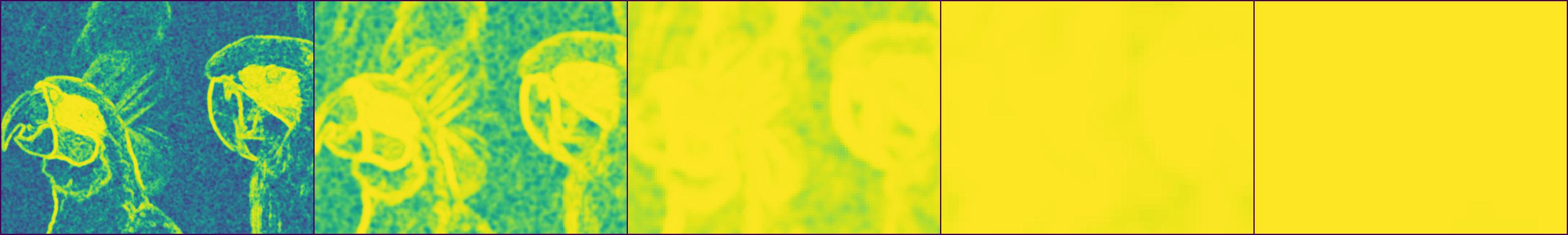}
        \raisebox{0.1\height}{\makebox[0.01\textwidth]{\rotatebox{90}{\makecell{\small Image B}}}}
        \includegraphics[width=.99\textwidth]{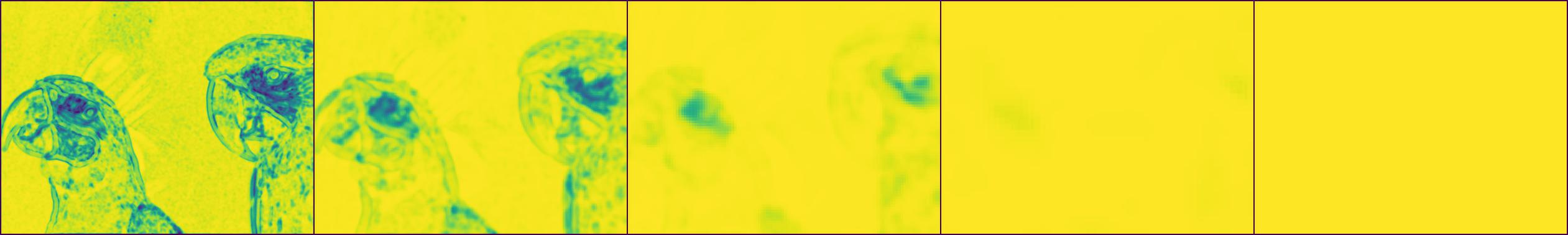}
    \end{subfigure}
    \vspace{2pt}
    \\
    \hdashrule[0.5ex]{1.02\linewidth}{0.5pt}{1.5mm}
    \\
    \begin{subfigure}[t]{.99\linewidth}
        \raisebox{0.1\height}{\makebox[0.01\textwidth]{\rotatebox{90}{\makecell{\small Image A}}}}
        \includegraphics[width=.99\textwidth]{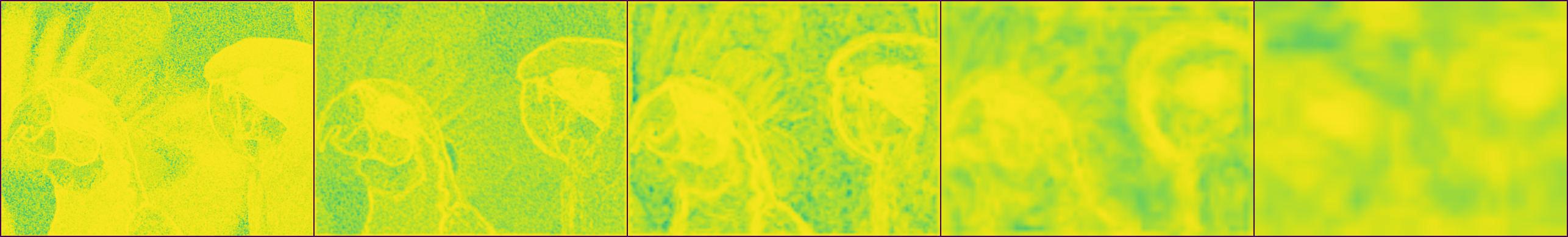}
        \raisebox{0.1\height}{\makebox[0.01\textwidth]{\rotatebox{90}{\makecell{\small Image B}}}}
        \includegraphics[width=.99\textwidth]{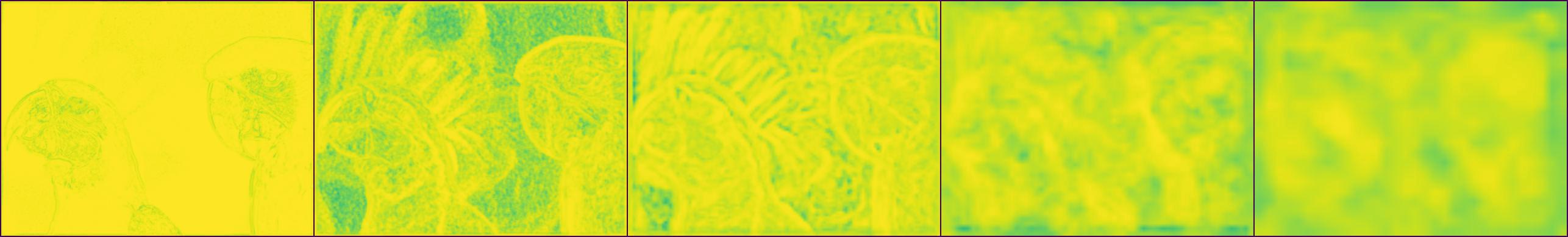}
    \end{subfigure}
    \vspace{2pt}
    \\
    \makebox[.01\linewidth]{}
    \makebox[.18\linewidth]{$384\times512$}
    \makebox[.19\linewidth]{$192\times256$}
    \makebox[.19\linewidth]{$96\times128$}
    \makebox[.19\linewidth]{$48\times64$}
    \makebox[.19\linewidth]{$24\times32$}
    \\
    \includegraphics[width=.99\linewidth]{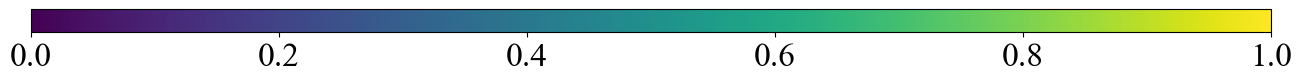}
    \caption{Multi-scale spatial quality maps ($H\times W$) of MS-SSIM (top two rows) and LPIPS (bottom two rows) with example images (Image A and Image B) from \cref{fig:intro_motivation}. Please zoom in for best view. \emph{Note: since LPIPS is lower better, we use (1 - LPIPS) here.}} \label{fig:local_map}
\end{figure}

\Cref{fig:local_map} shows the spatial quality maps of MS-SSIM and LPIPS before pooling for example images from \cref{fig:intro_motivation}. We have the following observations:
\begin{itemize}
    \item Both MS-SSIM and LPIPS appear to be distracted by the large background region in Image B, leading them to assign higher final scores to Image B. However, humans tend to focus more on the birds region and tend to prefer Image A.
    \item For these two cases, the high-level differences between Image A and Image B are small. MS-SSIM appears to have difficulties in extracting semantic features, and the pixel-level differences after downsampling are also small. On the other hand, the backbone network of LPIPS is capable of extracting high-level semantics, but it tends to lose distortion differences. Therefore, it can be challenging to determine which image is better based on high-level feature differences alone.
\end{itemize}

Based on these observations, we hypothesize that neither parallel nor bottom-up approaches can fully utilize multi-scale features. The parallel methods, such as MS-SSIM, have difficulties in extracting semantic representations. Conversely, for bottom-up approaches like LPIPS, although they can extract better semantic representations, they typically regress scores with different scale features independently, and therefore, are unable to focus on semantic regions as humans do.

\begin{figure}[!tbp]
    \centering
    \begin{subfigure}[t]{.99\linewidth}
        \centering
        \includegraphics[width=.99\linewidth]{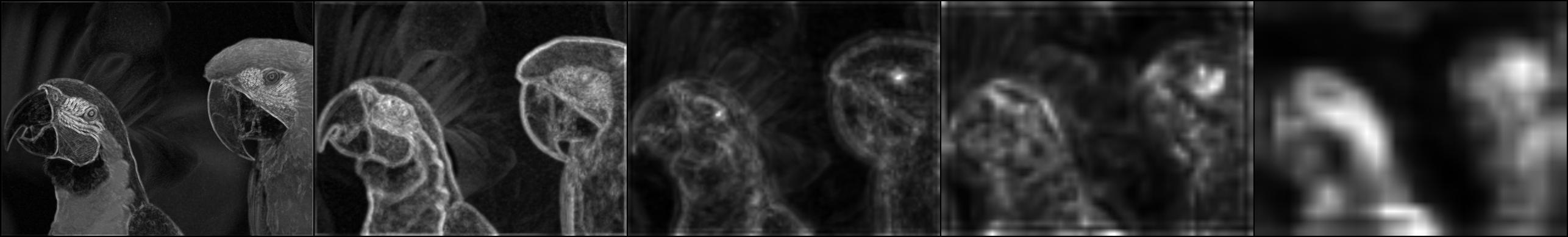}
        \caption{Example of multi-scale semantic activation maps in LPIPS from low-level to high-level. Please zoom in for best view.} \label{fig:semantic_maps}
    \end{subfigure}
    \begin{subfigure}[t]{.99\linewidth}
        \centering
        \includegraphics[width=.9\linewidth]{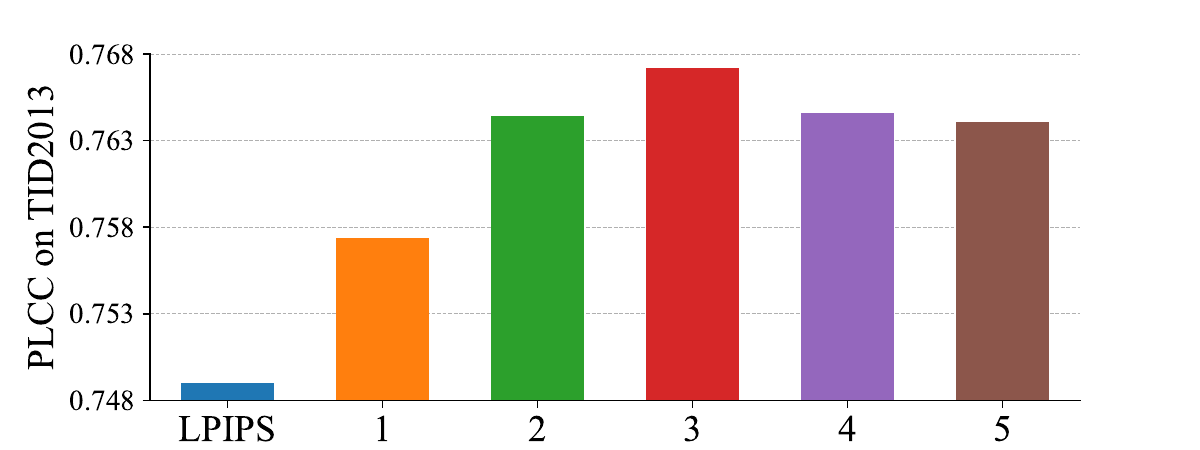}
        \caption{LPIPS+ using different layers as semantic weights.} \label{fig:lpips+}
    \end{subfigure}
    \caption{Empirical study of the LPIPS+ metric.  (a) The feature activation maps can be roughly taken as semantic weight maps; (b) The third layer semantic features bring the most improvement compared with original LPIPS.}
    \label{fig:vis_lpips+}
\end{figure}

\noindent \textbf{The LPIPS+ metric.} To verify our hypothesis, we explore a simple extension of LPIPS by replacing the average pooling with weighted average pooling, denoted as \textbf{LPIPS+}. We take the feature maps of reference images as rough estimations of semantic weights. As is known, features with higher activation values in neural networks usually correspond to semantic regions, as shown in \cref{fig:semantic_maps} for an example. Take reference features from $i$-th layer as $\F_i^r \in \R^{C_i \times H_i \times W_i}$, and the spatial quality map of $m$-th layer as $\S_m^r \in \R^{1 \times H_m \times W_m}$, LPIPS+ can be briefly formulated as follow:
\begin{equation}
   \text{LPIPS+} = \sum_m \frac{\sum \text{Resize}(\F_i^r) \odot \S_m^r}{\sum \text{Resize}(\F_i^r)},  
\end{equation}
where $\odot$ is element-wise multiplication, $\F_i^r$ is resized to the same shape as $\S_m^r$ using bilinear interpolation, and the summary dimension is omitted here for simplicity. From the examples in in \cref{fig:semantic_maps}, we can see that $\F_i^r$ in different layers display varying scales of semantic structures. As a result, we conducted an empirical study on TID2013 to evaluate the selection of semantic weight maps $\F_i^r$. The results, depicted in \cref{fig:lpips+}, show that all layers of semantic weight maps contribute to performance improvement, highlighting the importance of semantic information for multi-scale features. It is worth noting that each layer encompasses different scales of semantic structures, resulting in differing levels of performance enhancement. For LPIPS+, we selected $i=3$ based on our empirical findings. It is worth mentioning that LPIPS+ is an improved version of LPIPS that does not require additional training. 

The performance enhancements resulting from this simple extension have motivated us to develop a more robust framework that leverages the full potential of multi-scale features for IQA. To avoid the tedious and non-generalizable manual selection of multi-scale features across various datasets, we propose a heuristic top-down approach. This paradigm has proven to be effective in many different tasks, including object detection \cite{zheng2020cross} and semantic segmentation \cite{jing2019coarse}. In the following section, we provide details on our top-down framework.

\subsection{Architecture of Coarse-to-Fine Attention Network}

\begin{figure*}[htbp]
\centering
\includegraphics[width=\linewidth]{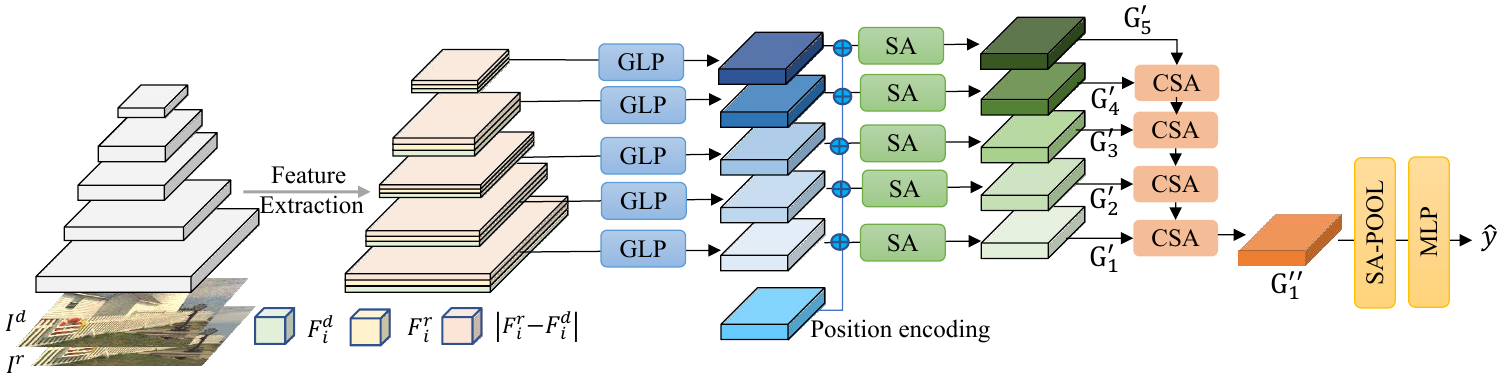}
\caption{Architecture overview of the proposed CFANet-FR. We use 5-scale features here same as previous works such as MS-SSIM and LPIPS.}
\label{fig:arch}
\end{figure*}

We have employed the top-down paradigm to develop the Coarse-to-Fine Attention Network (CFANet) to improve the utilization of multi-scale features for IQA, which can be applied to both FR and NR tasks. In this section, we focus on introducing the FR framework, as the NR framework is a simplified version. The pipeline of CFANet-FR is presented in \cref{fig:arch}. Given distortion-reference image pairs as input, we first extract their multi-scale features using a backbone network. Next, we employ gated local pooling (GLP) to reduce the multi-scale features to the same spatial size, which are then enhanced using self-attention (SA) blocks. Subsequently, we progressively apply cross-scale attention (CSA) blocks from high-level to low-level features. Finally, we pool the semantic-aware distortion features and regress them to the quality score through a multilayer perceptron (MLP). We provide a detailed explanation of each component below.

\subsubsection{Gated Local Pooling}

\begin{figure}[!t]
    \centering
    \includegraphics[width=.8\linewidth]{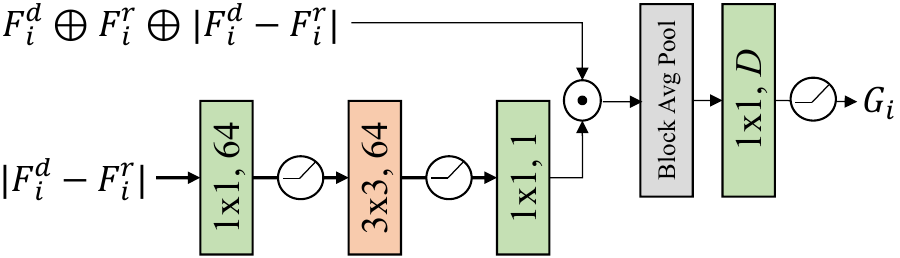}
    \caption{The GLP block comprises a mask branch and a feature branch. The mask branch is a bottleneck convolution block with an internal channel dimension of $64$. For FR datasets, we set the output dimension $D$ to $256$, and for NR datasets, we set it to $512$. All convolution layers are followed by the GELU activation function.
    } 
    \label{fig:glp}
\end{figure}

Denote input image pairs as $(I^d, I^r) \in \R^{3\times H \times W}$, the backbone features from block $i$ as $(\F_i^d, \F_i^r) \in \R^{C_i \times H_i \times W_i}$, where $H_i, W_i$ are height and width, $C_i$ is the channel dimension, $i\in\{1, 2, \ldots, n\}$ and $n=5$ for ResNet50. In general, low-level features are twice larger than their adjacent high-level features, and we have $H_i = H / 2^i$. Therefore, directly compute correlation between large matrix like $\F_1$ and $\F_2$ is too expensive. For simplicity and efficiency, we reduce $\F_i$ to the same shape as the highest level features $\F_n$. A na\"ive solution is simple window average pooling. However, this would fuse features inside local window and make the distortion feature less distinguishable. Instead, we propose to select the distortion related features before pooling through a gated convolution~\cite{gatedconv}, which has been proven to be useful in image inpainting. The problem here is how to calculate the gating mask. Notice that for FR task, the difference between $(\F_i^d, \F_i^r)$ is a strong clue for feature selection, we therefore formulate the gated convolution as 
\begin{equation}
    \F_i^{mask} = \sigma \left( \phi_i(|\F_i^d - \F_i^r|) \right) \cdot (\F_i^d \oplus \F_i^r \oplus |\F_i^d - \F_i^r|), \label{eq:gated_conv}
\end{equation}
where $\sigma$ is the sigmoid activation function that constrains the mask value to the range of $[0, 1]$, $\phi_i$ represents a bottleneck convolution block, and $\oplus$ denotes the concatenation operation. Please refer to \cref{fig:glp} for further details. For efficiency, we use a single-channel mask, \ie, $\phi_i(\cdot) \in \R^{1\times H_i \times W_i}$.

For the NR task, we use the same gated convolution formulation as follows:
\begin{equation}
\F_i^{mask} = \sigma \left(\phi_i (\F_i) \right) \cdot \text{ReLU} (W_f\F_i). \label{eq:nr_gated_conv}
\end{equation}
Subsequently, the masked feature $\F_i^{mask}$ undergoes window average pooling and a linear dimension reduction layer, producing features $\G_i \in \R^{D \times H_n \times W_n}$ for the following blocks, where $D$ denotes the reduced feature dimension. Our experiments show that our model can learn quality-aware masks and filter redundant features, as illustrated by the visualization of the gated mask.

\subsubsection{Attention Modules}

To help with the IQA task, we utilize the scaled dot-product attention \cite{vaswani2017attention} as the basis for our attention modules. Given triplets of feature vectors \emph{(query, key, value)}, the attention function first calculates similarities between the query ($\Q$) and key ($\K$) vectors and then outputs the weighted sum of values ($\V$). Suppose $\Q \in \R^{N_q\times d_k}, \K \in \R^{N_v\times d_k}, \V \in \R^{N_v\times d_v}$, the attention output is computed as
\begin{equation}
\text{Attn}(\Q, \K, \V) = \text{softmax}(\frac{\Q\K^T}{\sqrt{d_k}}) \V, \label{eq:attn}
\end{equation}
where $N_q$ and $N_v$ represent the number of feature vectors, and $d_k$ and $d_v$ indicate the feature dimension. We employ \cref{eq:attn} in various ways to aid the IQA task.

\paragraph{Self-attention}
After GLP, we obtain a set of features from different scales, denoted by $\{\G_1, \ldots, \G_n\} \in \R^{(H_n \times W_n) \times D}$. As the receptive field of low-level features is limited, we first enhance $\G_i$ with a self-attention block as follows:
\begin{equation}
\G_i' = \text{SA}(\G_i) = \text{Attn}(\G_i W_q , \G_i W_k, \G_i W_v) + \G_i,
\end{equation}
where $\G_i$ is projected onto $\Q, \K, \V$ through simple linear projection. Through the SA block, $\G_i'$ aggregates features from other positions to enhance $\G_i$. In \cite{tres2022wacv}, they concatenate the multi-scale features and use several transformer layers to regress the score, without considering the fact that different semantic regions hold different importance to humans. This approach does not allow for interaction between high-level semantic features and low-level distortion features, and thus cannot model such relationships. Our proposed cross-scale attention method addresses this issue in a straightforward manner.

\paragraph{Cross-scale Attention} 
Since the query feature $\Q$ in \cref{eq:attn} naturally serves as a guide when computing the output, our cross-attention is designed by simply generating the $\Q, \K, \V$ with features from different scales, \ie,  
\begin{align}
    \G_i'' &= \text{CSA}(\G_i', \G_{i+1}'') \nonumber \\
           &= \text{Attn}(W_q \G_{i+1}'', W_k \G_i', W_v \G_i') + \G_{i+1}'',
\end{align}
where $i\in \{1, \ldots, n - 1\}$, and $\G_n'' = \G_n'$. Intuitively speaking, the CSA block selects the most semantically relevant distortions in $\G_i'$ with high-level features $\G_{i+1}''$. The residual connection here serves as a simple fusion between features from different levels. The final output can be obtained by progressively applying CSA as
\begin{equation}
    \G''_1 = \text{CSA}\bigl( \ldots \text{CSA}\left(\G_{n-2}', \text{CSA}(\G_{n-1}', \G_n')\right) \bigr).
\end{equation}

\subsubsection{Unified position encoding} In transformers, position encoding is crucial to inject awareness of feature positions in \cref{eq:attn}. In our CSA blocks, position information is also important as another clue for cross-scale feature query. In \cite{ke2021musiq}, Ke \etal designed a hash-based 2D spatial embedding for multi-scale inputs. In our framework, since the multi-scale features $\G_i$ have the same shape after GLP, we simply add the same learnable position encoding to all $\G_i$, as shown in \cref{fig:arch}. This unified position encoding enables CSA to better match features from different scales.

\subsubsection{Score Regression} The final scores are obtained using the final features $\G_1''$ as follows:
\begin{equation}
\hat{y} = \text{MLP} \bigl ( \text{SA-Pool}(\mathbf{\G_1''}) \bigr), \label{eq:mlp}
\end{equation}
where SA-Pool is a self-attention block followed by average pooling. The SA block is added to better assemble features from all positions. When predicting score distributions, we have $\hat{p}=\text{softmax}(\hat{y})$.

\subsection{Loss Functions}

Since different datasets have different kinds of labels, we need different losses for them, which are detailed below:

\subsubsection{MOS labeled datasets} For these datasets, we first normalize the MOS scores to $[0, 1]$ and then use the MSE loss. 

\subsubsection{MOS distribution labels} For datasets that are labeled with score distributions, such as the AVA dataset \cite{murray2012ava}, we predict the distribution and use the Earth Mover's Distance (EMD) loss proposed by \cite{talebi2018nima}.

\subsubsection{2AFC datasets} Some recent large scale datasets, such as PieAPP \cite{Prashnani_2018_PieAPP} and BAPPS \cite{zhang2018lpips} are labeled with preference through 2AFC (two-alternative force choice\footnote{The subjects need to choose a better one given two candidates.}) rather than single MOS label. Given triplet pairs, a reference image with two distorted images denoted as $(I_r, I_A, I_B)$, the datasets provide the probability of subject preference to one of $I_A$ and $I_B$. Following the same practice of \cite{Prashnani_2018_PieAPP}, we first learn the perceptual error scores for $I_A$ and $I_B$ with the network separately, \ie,
\begin{equation}
    \hat{y}_A = \text{CFANet}(I_r, I_A), \quad 
    \hat{y}_B = \text{CFANet}(I_r, I_B).
\end{equation}
Then, $\hat{y}_A$ and $\hat{y}_B$ are used to compute the preference probability of $I_A$ over $I_B$ with the Bradley-Terry (BT) sigmoid model~\cite{bradley1952rank} as follows,
\begin{equation}
    \hat{p}_{AB} = \frac{1}{1 + e^{\hat{y}_A - \hat{y}_B}}. \label{eq:bt_model}
\end{equation}
The common MSE is finally used as the loss function:
\begin{equation}
    L_{2AFC} (\hat{y}_A, \hat{y}_B, p_{AB}) = \frac{1}{N} \sum_{i=1}^N\| \hat{p}_{AB} - p_{AB}\|^2.
\end{equation}


\section{Experiments}

\subsection{Implementation Details} 

\subsubsection{Datasets} 

\begin{table*}[t]
\centering
\caption{FR and NR IQA Datasets used for training and evaluation.} \label{tab:datasets}
\renewcommand*{\arraystretch}{1.2}
\resizebox{\linewidth}{!}{
\begin{tabular}{clccccccccc}
\toprule
Type & Dataset &  \# Ref & \# Dist & Dist Type. & \# Rating & Split & \makecell[c]{Original size \\ $W\times H$}& \makecell[c]{Resize \\ (shorter side)} & \makecell[c]{Train size \\ (cropped patch)} \\ \midrule
\multirow{7}{*}{FR} & LIVE & 29 & 779 & Synthetic & 25k & 6:2:2 & $768\times512$ (typical) & --- & $384\times384$ \\  
& CSIQ & 30 & 866 & Synthetic & 5k & 6:2:2 & $512\times512$ & --- & $384\times384$ \\ 
& TID2013 & 25 & 3,000 & Synthetic & 524k & 6:2:2 & $512\times384$ & --- & $384\times384$  \\ 
& KADID-10k & 81 & 10.1k & Synthetic & 30.4k & 6:2:2 & $512\times384$ & --- & $384\times384$ \\ 
& PieAPP & 200 & 20k & Synthetic & 2.3M & Official & $256\times256$ & --- & $224\times224$ \\ 
& BAPPS & -- & 187.7k & Syth.+alg. & 484k & Official & $500\times500$ & --- & $384\times384$ \\ 
& PIPAL & 250 & 29k & Syth.+alg. & 1.13M & Official & $288\times288$ & --- & $224\times224$ \\ 
\midrule
\multirow{5}{*}{NR} & CLIVE & \multirow{5}{*}{--} & 1.2k & Authentic & 350k & 8:2 & $500\times500$ & --- & $384\times384$ \\
& KonIQ-10k & & 10k & Authentic & 1.2M & 8:2 &  $512\times384$ & --- & $384\times384$ \\
& SPAQ & & 11k & Authentic & -- & 8:2 & 4K (typical) & 448 & $384\times384$\\
& AVA & & 250k & Aesthetic & 53M & Official & $< 800$ & $384 \sim 416$ & $384\times384$ \\
& FLIVE & & 160k & Auth.+Aest. & 3.9M & Official & Train$<640$ $\mid$ Test$>640$ & $384 \sim 416$ & $384\times384$ \\
\bottomrule
\end{tabular}
}
\end{table*}

As shown in \cref{tab:datasets}, we conduct experiments on several public benchmarks. For FR datasets, we have LIVE~\cite{sheikh2006liveiqa}, CSIQ~\cite{larson2010csiq}, TID2013~\cite{ponomarenko2013tid2013}, KADID-10k~\cite{kadid10k}, PieAPP~\cite{Prashnani_2018_PieAPP}, BAPPS~\cite{zhang2018lpips} and PIPAL~\cite{pipal}. For NR datasets, we have got CLIVE~\cite{livechallenge}, KonIQ-10k~\cite{koniq10k}, SPAQ~\cite{fang2020spaq}, FLIVE~\cite{flivepaq2piq} and AVA~\cite{murray2012ava}. We use the official train/val/test splits if available, otherwise, we randomly split it 10 times and report the mean and variance. For FR datasets, the split is based on reference images to avoid content overlapping.

\subsubsection{Performance Evaluation} 
We applied two commonly used metrics: the Pearson linear correlation coefficient (PLCC) and the Spearman’s rank-order correlation coefficient (SRCC). PLCC measures the linear correlation between predicted scores ($\hat{y}$) and ground truth labels ($y$), while SRCC assesses rank correlation. 
The same as \cite{2020dbcnn,dists2020}, we fitted a 4-parameter logistic function to the predicted scores before calculating PLCC:
\begin{equation}
    \hat{y}' = \frac{\beta_1 - \beta_2}{1 + \exp(-(\hat{y} - \beta_3) / |\beta_4|)} + \beta_2,
\end{equation}
where $\{\beta_i|i=1, 2, 3, 4\}$ are fitted with least square losses between $\hat{y}'$ and GT labels $y$, and are initialized with $\beta_1 = \max(y), \beta_2 = \min(y), \beta_3 = \mu(\hat{y}), \beta_4=\sigma(\hat{y}) / 4$. Here, $\sigma(\cdot)$ is the standard variation.  

\subsubsection{Training Details}  
We use ResNet50, pretrained on ImageNet~\cite{imagenet}, as the backbone for most of our experiments. As is common in domain transfer, we fix the batch normalization layers and finetune the other parameters. We use data augmentation operators that do not affect image quality, such as random crop and horizontal/vertical flip. We use the AdamW optimizer with a weight decay of $10^{-5}$ for all experiments. The initial learning rate ($lr$) is set to $10^{-4}$ for FR datasets and $3\times10^{-5}$ for NR datasets. We use a cosine annealing scheduler with $T_{max}=50, \eta_{min}=0, \eta_{max}=lr$, following previous works~\cite{cheon2021iqt,lao2022ahiq}. The total number of training epochs is 200, and we use early stopping based on validation performance to reduce training time. Our model is implemented using PyTorch and trained on an NVIDIA V100 GPU.

We keep the training settings, including network hyperparameters and optimizer settings, consistent across different FR and NR benchmarks. However, due to differences in image sizes across datasets, we have to resize the images to an appropriate size for training the network. As shown in \cref{tab:datasets}, images from three datasets, SPAQ, AVA, and FLIVE, need to be resized. To preserve image quality, we maintain aspect ratio during resizing and \cref{tab:datasets} shows the size of the shorter side after resize. For AVA and FLIVE, we randomly set the shorter side between 384 and 416 as a data augmentation strategy.

\subsection{Visualization of Attention Maps} 

\begin{figure*}[pt]
    \centering
    \begin{subfigure}[t]{.99\linewidth}
    \makebox[.19\linewidth]{Distorted Image}
    \makebox[.19\linewidth]{$\phi_4(\cdot)$}
    \makebox[.19\linewidth]{$\phi_3(\cdot)$}
    \makebox[.19\linewidth]{$\phi_2(\cdot)$}
    \makebox[.19\linewidth]{$\phi_1(\cdot)$}
    \\
    \includegraphics[width=.197\linewidth]{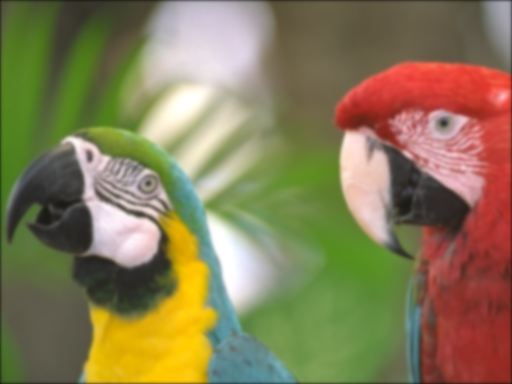}  \includegraphics[width=.79\linewidth]{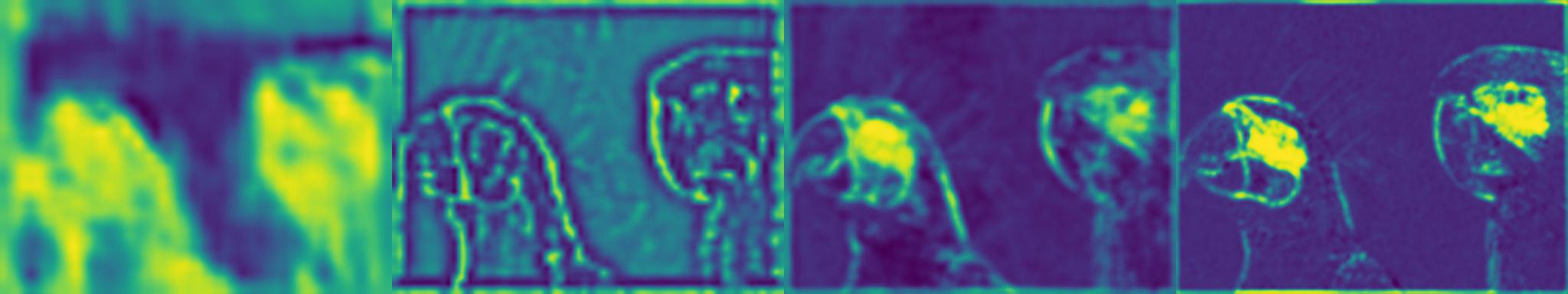} 
    \\
    \includegraphics[width=.197\linewidth]{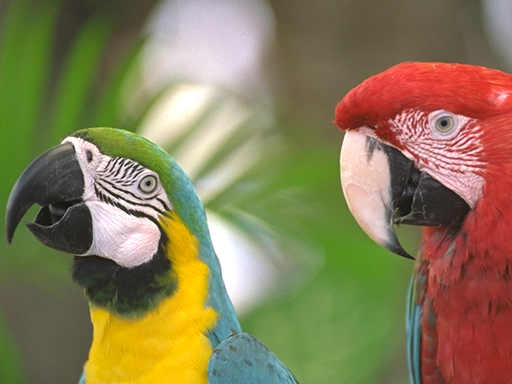}  \includegraphics[width=.79\linewidth]{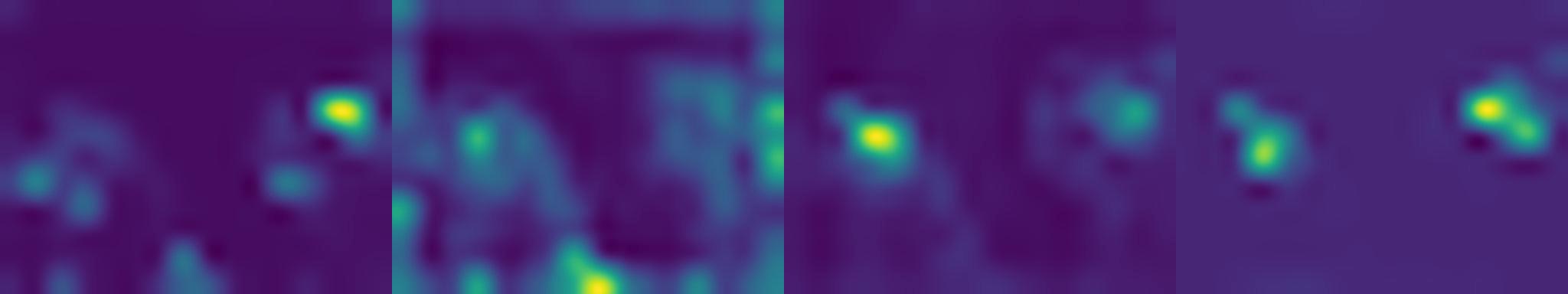} 
    \\
    \makebox[.19\linewidth]{Reference Image}
    \makebox[.19\linewidth]{$\text{CSA}(\F_5 \rightarrow \F_4$)}
    \makebox[.19\linewidth]{$\text{CSA}(\F_4 \rightarrow \F_3)$}
    \makebox[.19\linewidth]{$\text{CSA}(\F_3 \rightarrow \F_2$)}
    \makebox[.19\linewidth]{$\text{CSA}(\F_2 \rightarrow \F_1$)}
    \caption{Example with ``gaussian blur'' distortion.}
    \label{fig:id_blur}
    \end{subfigure}
    \\ \vspace{0.5em}
    \begin{subfigure}[t]{.99\linewidth}
    \makebox[.19\linewidth]{Distorted Image}
    \makebox[.19\linewidth]{$\phi_4(\cdot)$}
    \makebox[.19\linewidth]{$\phi_3(\cdot)$}
    \makebox[.19\linewidth]{$\phi_2(\cdot)$}
    \makebox[.19\linewidth]{$\phi_1(\cdot)$}
    \\
    \includegraphics[width=.197\linewidth]{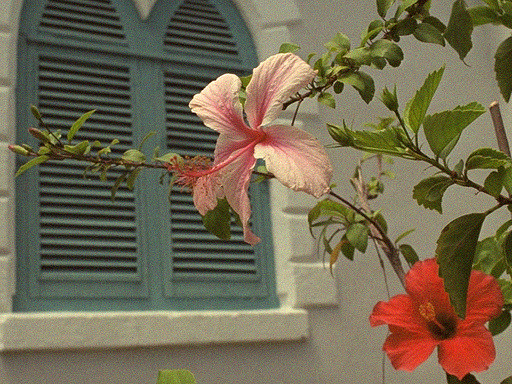}  \includegraphics[width=.79\linewidth]{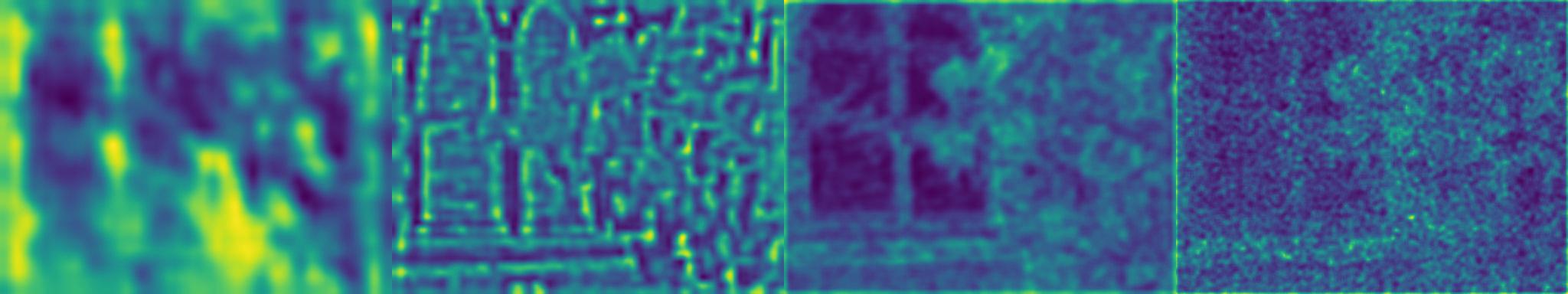} 
    \\
    \includegraphics[width=.197\linewidth]{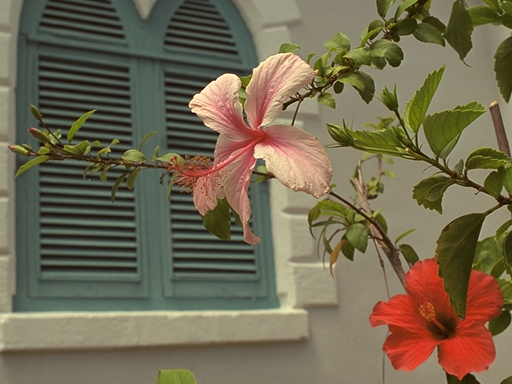}  \includegraphics[width=.79\linewidth]{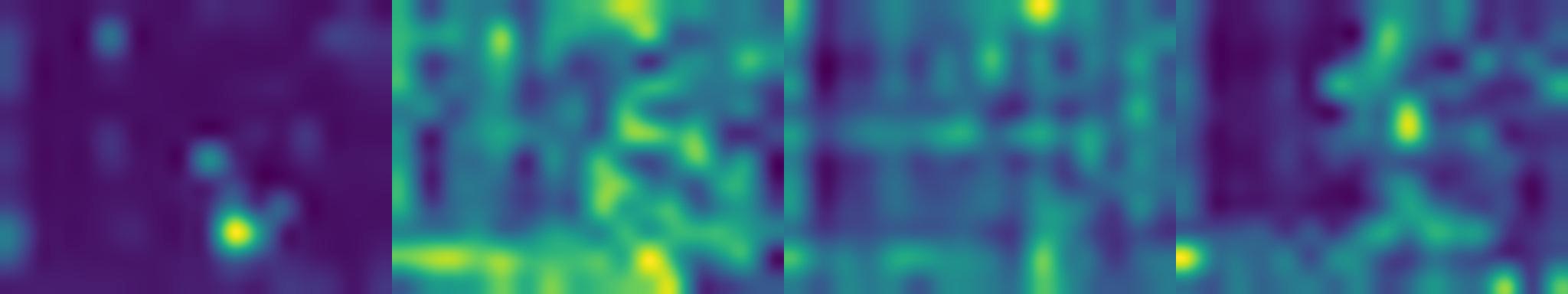} 
    \\
    \makebox[.19\linewidth]{Reference Image}
    \makebox[.19\linewidth]{$\text{CSA}(\F_5 \rightarrow \F_4$)}
    \makebox[.19\linewidth]{$\text{CSA}(\F_4 \rightarrow \F_3)$}
    \makebox[.19\linewidth]{$\text{CSA}(\F_3 \rightarrow \F_2$)}
    \makebox[.19\linewidth]{$\text{CSA}(\F_2 \rightarrow \F_1$)}
    \caption{Example with ``high frequency noise'' distortion.}
    \label{fig:tid_hfnoise}
    \end{subfigure}
    \\ \vspace{0.5em}
    \begin{subfigure}[t]{.99\linewidth}
    \makebox[.19\linewidth]{Distorted Image}
    \makebox[.19\linewidth]{$\phi_4(\cdot)$}
    \makebox[.19\linewidth]{$\phi_3(\cdot)$}
    \makebox[.19\linewidth]{$\phi_2(\cdot)$}
    \makebox[.19\linewidth]{$\phi_1(\cdot)$}
    \\
    \includegraphics[width=.197\linewidth]{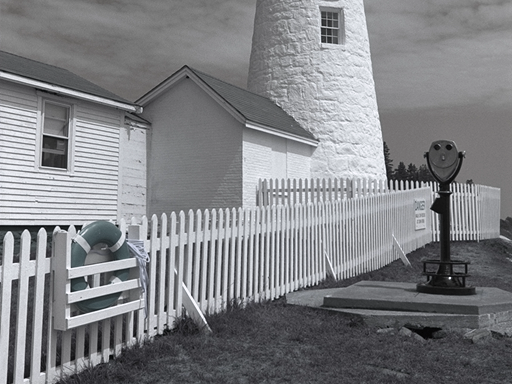}  \includegraphics[width=.79\linewidth]{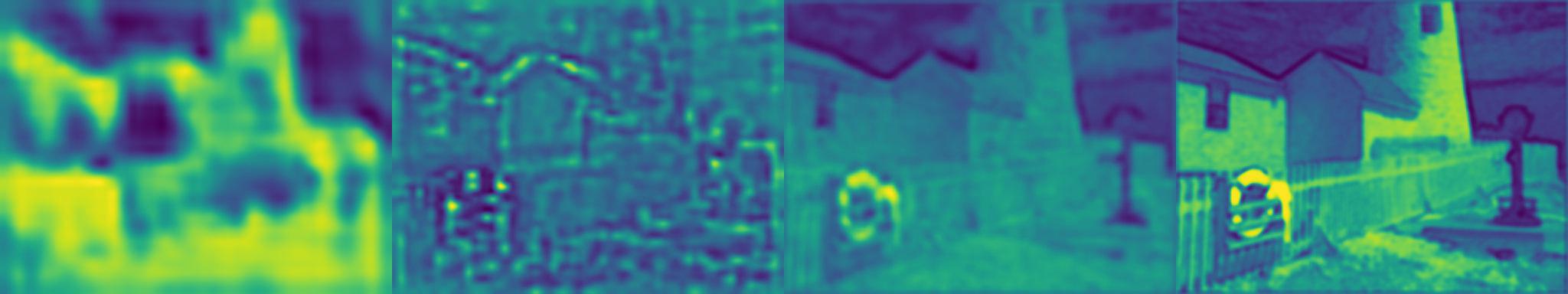} 
    \\
    \includegraphics[width=.197\linewidth]{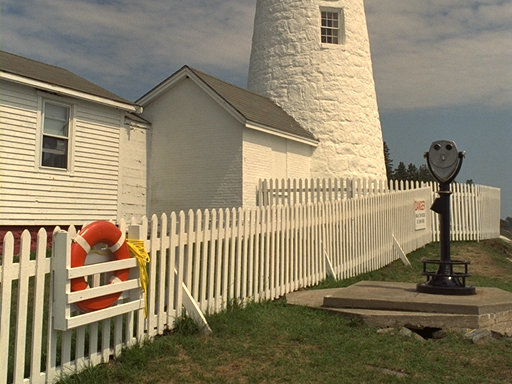}  \includegraphics[width=.79\linewidth]{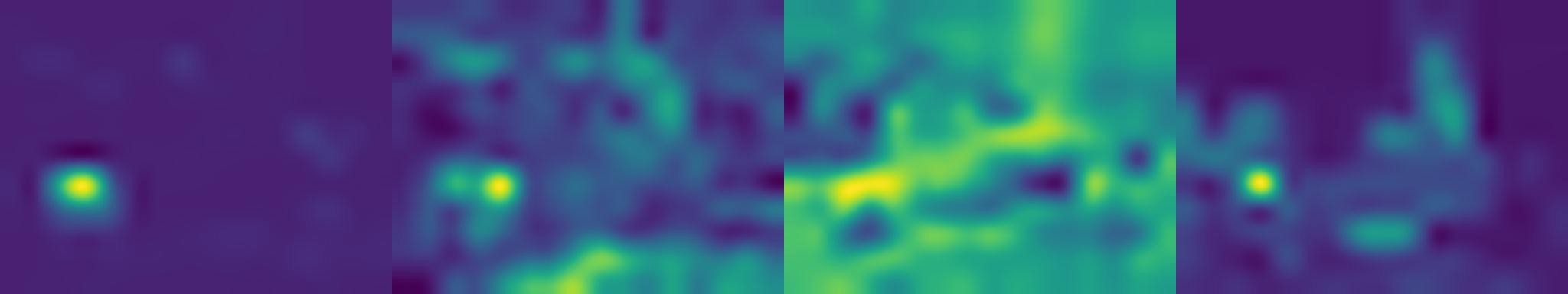} 
    \\
    \makebox[.19\linewidth]{Reference Image}
    \makebox[.19\linewidth]{$\text{CSA}(\F_5 \rightarrow \F_4$)}
    \makebox[.19\linewidth]{$\text{CSA}(\F_4 \rightarrow \F_3)$}
    \makebox[.19\linewidth]{$\text{CSA}(\F_3 \rightarrow \F_2$)}
    \makebox[.19\linewidth]{$\text{CSA}(\F_2 \rightarrow \F_1$)}
    \caption{Example with ``change of color saturation''.}
    \label{fig:tid_color}
    \end{subfigure}
    \caption{Attention visualization with different distortion types from TID2013 dataset. First row: GLP mask, $\phi_i(\cdot)$ in \cref{eq:gated_conv,eq:nr_gated_conv}; Second row: CSA attention weights.}
    \label{fig:attn_vis}
\end{figure*}

In this part, we visualize attention maps to show how CFANet works in a top-down manner. CFANet has two types of attention maps: i) the distortion attention masks learned in GLP and ii) the cross-scale attention maps learned in CSA blocks. The former filters redundant information and reduces the spatial size of feature maps, while the latter enables semantic propagation from coarse to fine. \Cref{fig:attn_vis} shows the visualization of the learned masks in GLP blocks for multi-scale features $\F_1, \cdots, \F_4$ and the cross-scale attention weights from $\F_{i+1}$ to $\F_i$ in CSA blocks. Examples of three different distortions, \ie, ``gaussian blur", ``high frequency noise", and ``change of color saturation", are presented. 

We can observe that GLP blocks can selectively identify distortion-related features at different scales for different types of distortions, especially in $\F_1$. The CSA attention maps show that the model gradually focuses on semantic regions in a coarse-to-fine manner. For example, in \cref{fig:id_blur} (the Image B in \cref{fig:intro_motivation}), the network is not distracted by the large background regions and is able to focus on the birds. This explains why CFANet makes consistent judgements with humans in the case in \cref{fig:intro_motivation}. Similar observations can be found in \cref{fig:tid_hfnoise} and \cref{fig:tid_color}, which prove that CFANet is robust to different types of distortions. These observations demonstrate that CFANet effectively extracts semantically important distortion features. 

\subsection{Comparison with FR Methods}

\begin{table*}[t]
\centering
\caption{Quantitative comparison with related works on public \textbf{FR benchmarks}, including the traditional LIVE, CSIQ, TID2013 with MOS labels, and recent large scale datasets PieAPP, PIPAL with 2AFC labels. The best and second results are colored in \best{red} and \second{blue}, and ``-'' indicates the score is not available or not applicable.} \label{tab:fr_intra}
\resizebox{\linewidth}{!}{
\begin{tabular}{lcccccccccc}
\toprule
& \multicolumn{2}{c}{LIVE\cite{sheikh2006liveiqa}} & \multicolumn{2}{c}{CSIQ\cite{larson2010csiq}} & \multicolumn{2}{c}{TID2013\cite{ponomarenko2013tid2013}} & \multicolumn{2}{c}{\textbf{PieAPP\cite{Prashnani_2018_PieAPP}}} & \multicolumn{2}{c}{\textbf{PIPAL\cite{pipal}}}
\\
\cmidrule(lr){2-3}
\cmidrule(lr){4-5}
\cmidrule(lr){6-7}
\cmidrule(lr){8-9}
\cmidrule(lr){10-11}
Method & PLCC & SRCC & PLCC & SRCC & PLCC & SRCC & PLCC & SRCC & PLCC & SRCC 
\\ \midrule
PSNR & 0.865 & 0.873 & 0.819 & 0.810 & 0.677 & 0.687 & 0.135 & 0.219 & 0.277 & 0.249 \\ 
SSIM~\cite{ssim2004} & 0.937 & 0.948 & 0.852 & 0.865 & 0.777 & 0.727 & 0.245 & 0.316 & 0.391 & 0.361 \\ 
MS-SSIM~\cite{msssim2003} & 0.940 & 0.951 & 0.889 & 0.906 & 0.830 & 0.786 & 0.051 & 0.321 & 0.163 & 0.369 \\ 
VIF~\cite{sheikh2006vif} & 0.960 & 0.964 & 0.913 & 0.911 & 0.771 & 0.677 & 0.250 & 0.212 & 0.479 & 0.397 \\
FSIMc~\cite{zhang2011fsim} & 0.961 & 0.965 & 0.919 & 0.931 & 0.877 & 0.851 & 0.481 & 0.378 & 0.571 & 0.504 \\ 
MAD~\cite{larson2010mad} & 0.968 & 0.967 & 0.950 & 0.947 & 0.827 & 0.781 & 0.231 & 0.304 & 0.580 & 0.543 \\
GMSD~\cite{xue2013gmsd} & 0.957 & 0.960 & 0.945 & 0.950 & 0.855 & 0.804 & 0.242 & 0.297 & 0.608 & 0.537 \\
VSI~\cite{zhang2014vsi} & 0.948 & 0.952 & 0.928 & 0.942 & 0.900 & 0.897 & 0.364 & 0.361 & 0.517 & 0.458 \\ 
NLPD~\cite{laparra2016nlpd} & 0.932 & 0.937 & 0.923 & 0.932 & 0.839 & 0.800 & 0.360 & 0.245 & 0.401 & 0.355 \\
\midrule
DeepQA~\cite{kim2017deepqa} & 0.982 & \second{0.981} & 0.965 & 0.961 & 0.947 & 0.939 & 0.172 & 0.252 & - & - \\
WaDIQaM-FR~\cite{bosse2017wadiqam} & 0.980 & 0.970 & - & - & 0.946 & 0.940 & 0.439 & 0.352 & 0.548 & 0.553 \\
PieAPP~\cite{Prashnani_2018_PieAPP} & 0.986 & 0.977 & 0.975 & 0.973 & 0.946 & 0.945 & \second{0.842} & 0.831 & 0.597 & 0.607  \\
LPIPS-VGG~\cite{zhang2018lpips} & 0.978 & 0.972 & 0.970 & 0.967 & 0.944 & 0.936 & 0.654 & 0.641 & 0.633 & 0.595 \\
DISTS~\cite{dists2020} & 0.980 & 0.975 & 0.973 & 0.965 & 0.947 & 0.943 & 0.725 & 0.693 & 0.687 & 0.655 \\
JND-SalCAR~\cite{seo2020jndsalsar} & \second{0.987} & \best{0.984} & 0.977 & \second{0.976} & 0.956 & 0.949 & - & - & - & -\\
IQT~\cite{cheon2021iqt} & - & - & - & - & - & - & 0.829 & 0.822 & 0.790 & \second{0.799}  \\ 
AHIQ~\cite{lao2022ahiq} & \best{0.989} & \best{0.984} & \second{0.978} & 0.975 & \best{0.968} & \best{0.962} & 0.840 & \second{0.838} & \second{0.823} & \best{0.813} \\
\midrule
\ours (CFANet-ResNet50) & 0.984 & \best{0.984} & \best{0.980} & \best{0.978} & \second{0.958} & \second{0.954} & \best{0.849} & \best{0.841} & \best{0.830} & \best{0.813} \\ 
std & $\pm$0.003 & $\pm$0.003 & $\pm$0.003 & $\pm$0.002 & $\pm$0.011 & $\pm$0.012 & - & - & - & -\\
\bottomrule
\end{tabular}
}
\end{table*}
\begin{table*}[t]
\centering
\caption{Comparison of cross-dataset performance on public benchmarks. } \label{tab:fr_cross}
\resizebox{\linewidth}{!}{
\begin{tabular}{lcccccccccccc}
\toprule
\textbf{Train dataset} & \multicolumn{6}{c}{KADID-10k} & \multicolumn{6}{c}{PIPAL} \\ 
\cmidrule(rl){2-7}
\cmidrule(rl){8-13}
\textbf{Test dataset} & \multicolumn{2}{c}{LIVE} & \multicolumn{2}{c}{CSIQ} & \multicolumn{2}{c}{TID2013} & \multicolumn{2}{c}{LIVE} & \multicolumn{2}{c}{CSIQ} & \multicolumn{2}{c}{TID2013} \\
\cmidrule(rl){2-3}
\cmidrule(rl){4-5}
\cmidrule(rl){6-7}
\cmidrule(rl){8-9}
\cmidrule(rl){10-11}
\cmidrule(rl){12-13}
Method & PLCC & SRCC & PLCC & SRCC & PLCC & SRCC & PLCC & SRCC & PLCC & SRCC & PLCC & SRCC 
\\ \midrule
WaDIQaM-FR~\cite{bosse2017wadiqam} & 0.940 & 0.947 & 0.901 & 0.909 & 0.834 & 0.831 & 0.895 & 0.899 & 0.834 & 0.822 & 786 & 0.739 \\
PieAPP~\cite{Prashnani_2018_PieAPP} & 0.908 & 0.919 & 0.877 & 0.892 & 0.859 & 0.876 & - & - & - & - & - & - \\
LPIPS-VGG~\cite{zhang2018lpips} & 0.934 & 0.932 & 0.896 & 0.876 & 0.749 & 0.670 & 0.901 & 0.893 & 0.857 & 0.858 & 0.790 & 0.760 \\
DISTS~\cite{dists2020} & \second{0.954} & 0.954 & 0.928 & 0.929 & 0.855 & 0.830 & 0.906 & 0.915 & \second{0.862} & 0.859 & 0.803 & \second{0.765} \\
AHIQ~\cite{lao2022ahiq} & 0.952 & \second{0.970} & \second{0.955} & \second{0.951} & \second{0.899} & \second{0.901} & \second{0.911} & \second{0.920} & 0.861 & \second{0.865} & \second{0.804} & 0.763 \\
\midrule
\ours (Resnet50) & \best{0.957} & \best{0.974} & \best{0.963} & \best{0.969}  & \best{0.916} & \best{0.915} & \best{0.913} & \best{0.939} & \best{0.908} & \best{0.908} & \best{0.846} & \best{0.816} \\ 
\bottomrule
\end{tabular}
}
\end{table*}

\begin{figure}[t]
    \centering
    \includegraphics[width=\linewidth]{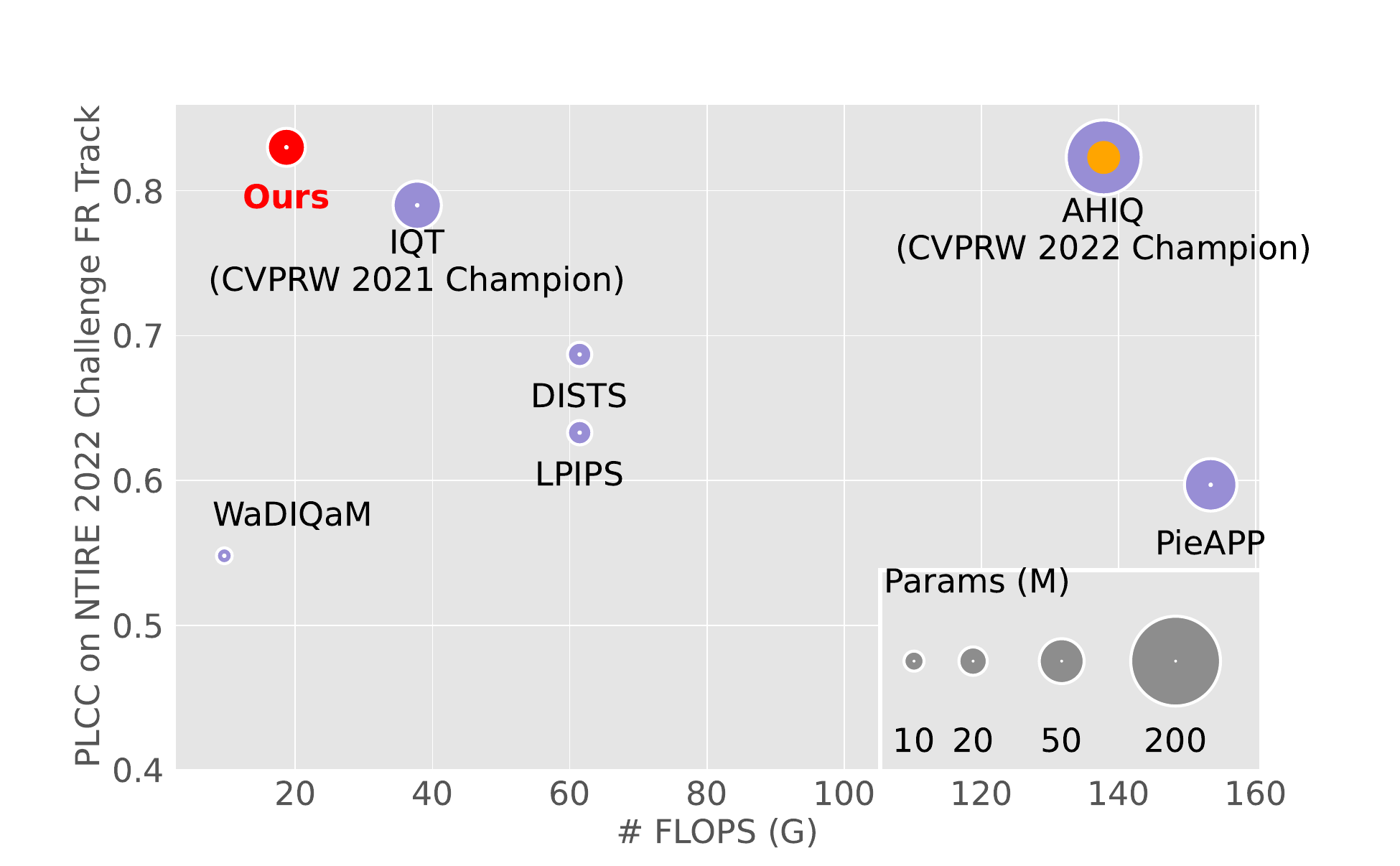}
    \caption{Computational cost (FLOPS) vs. Performance (PLCC) on NTIRE IQA Challenge 2022 FR Track. Our model achieves the best performance with only ${\sim}13\%$ FLOPS as previous state-of-the-art AHIQ. \emph{Note: The input image size is $3\times224\times224$. The number of parameters is indicated by the circle radius. For AHIQ, the backbone is fixed and the number of trainable parameters is indicated by the orange circle.}}
    \label{fig:computation}
\end{figure}

To demonstrate the superiority of the top-down approach, we compare our proposed CFANet to various traditional and deep learning methods using FR benchmarks (see \cref{tab:datasets}). Our evaluations include both intra-dataset and cross-dataset experiments. Additionally, we compare our results to those of the widely recognized LPIPS using the same experimental setup.

\subsubsection{Intra-dataset results of public benchmarks} We conducted intra-dataset experiments on five benchmarks, namely LIVE, CSIQ, TID2013, PieAPP, and PIPAL. The first three datasets are small synthetic datasets labeled with MOS scores, while the latter two are much larger datasets labeled through 2AFC and contain a wider variety of distortion types. The results are presented in \cref{tab:fr_intra}. As we can see, both traditional and deep learning methods perform well on the easier conventional benchmarks, LIVE, CSIQ, and TID2013, which only contain a few types of synthetic distortions. In particular, the proposed CFANet performs as well as AHIQ and demonstrates remarkable performance. It's important to note that performance on these three datasets can vary significantly due to different splits, especially for TID2013 according to the variance.

Regarding the larger-scale datasets, PieAPP and PIPAL, our CFANet outperforms all previous methods, including the AHIQ with a heavy transformer backbone. Notably, our CFANet achieves this with a simple ResNet50 backbone, demonstrating the remarkable effectiveness of the proposed top-down framework for IQA.

\subsubsection{Cross dataset experiments} Furthermore, CFANet exhibits significantly better generalization abilities with fewer parameters, as reported in \cref{tab:fr_cross}. With the current largest dataset, PIPAL, containing only 29k pairs\footnote{Due to ambiguities in human perception, one image pair usually requires dozens of annotations to obtain the final MOS, making it expensive to build large-scale datasets for IQA.}, larger models also face the issue of overfitting. Comparing the results in \cref{tab:fr_intra} and \cref{tab:fr_cross}, we can observe that the performance gaps of AHIQ on LIVE, CSIQ, and TID2013 are much larger than those of CFANet, demonstrating that the simpler CFANet is more robust across different datasets.

\subsubsection{Comparison of computation complexity} 
Figure \ref{fig:computation} presents an intuitive comparison of the computational expenses of recent deep learning-based FR methods. It is evident that CFANet exhibits the best performance with only approximately 13\% FLOPS and around 1/7 of AHIQ's parameters. While earlier works with simpler architectures, such as WaDIQaM, are more efficient, their performance is notably inferior. With the aid of the efficient ResNet50 backbone, CFANet is also more efficient than LPIPS. In terms of inference time, methods with CNN backbones, including CFANet, are comparable and nearly twice as fast as transformer-based approaches like AHIQ. In summary, CFANet strikes the best balance between performance and computational complexity.

\subsubsection{Comparison on BAPPS dataset} BAPPS~\cite{zhang2018lpips} is a 2AFC FR dataset proposed by the widely recognized LPIPS. Because its evaluation protocol differs from other mainstream datasets, we provide a separate comparison experiment on BAPPS in this section. The validation set of BAPPS only has binary preference labels, so we cannot calculate PLCC and SRCC scores. Instead, LPIPS uses the consistency between model preference and human judgment to calculate the final score, which is defined as follows:
\begin{align}
    \text{Score} &= \1(\hat{y}_A < \hat{y}_B) \1(p_A < p_B) \nonumber \\ 
    & + \1(\hat{y}_A > \hat{y}_B) \1(p_A > p_B) + 0.5\1(\hat{y}_A = \hat{y}_B).
\end{align}
This score only measures the binary preference judgements rather than exact probability values.

The comparison of CFANet and other methods on the 2AFC test set of BAPPS is shown in \cref{tab:bapps}. We can observe that the proposed CFANet achieves the best performance on both synthetic and real algorithmic distortions, outperforming previous approaches by a large margin. Our results are very close to human judgments, especially on synthetic distortions. In addition, we also tested the proposed LPIPS+. The results show that LPIPS+ outperforms LPIPS in almost all sub-tasks, further proving the effectiveness of semantic guidance for IQA.   

\begin{table*}[t]
    \caption{Performance comparison on the 2AFC test set of BAPPS (2AFC score, higher is better). Note: the results ``All'' are simply calculated as the mean value of corresponding sub-terms same as \cite{zhang2018lpips}.}
    \label{tab:bapps}
    \centering
    \resizebox{\linewidth}{!}{
    \begin{tabular}{l*{9}{c}}
    \toprule
    \multirow{3}{*}{Method} & \multicolumn{3}{c}{Synthetic distortions} & \multicolumn{5}{c}{Distortions by real algorithms} & \multirow{3}{*}{All} \\
    \cmidrule(lr){2-4}
    \cmidrule(lr){5-9}
    & Traditional & CNN-based & All & \makecell[c]{Super \\ resolution} & \makecell[c]{Video \\ deblurring} & Colorization & \makecell[c]{Frame \\ interpolation} & All & 
    \\ \midrule
    \rowcolor{lightcyan} Human & 0.808 & 0.844 & 0.826 & 0.734 & 0.671 & 0.688 & 0.686 & 0.695 & 0.739 \\ \midrule
    PSNR & 0.573 & 0.801 & 0.687 & 0.642 & 0.590 & 0.624 & 0.543 & 0.600 & 0.629 \\ 
    SSIM &  0.605 & 0.806 & 0.705 & 0.647 & 0.589 & 0.624 & 0.573 & 0.608 & 0.641 \\ 
    MS-SSIM & 0.585 & 0.768 & 0.676 & 0.638 & 0.589 & 0.524 & 0.572 & 0.581 & 0.613\\
    VSI & 0.630 & 0.818 & 0.724 & 0.668 & 0.592 & 0.597 & 0.568 & 0.606 & 0.646 \\
    MAD &     0.598 & 0.770 & 0.684 & 0.655 & 0.593 & 0.490 & 0.581 & 0.580 & 0.615 \\
    VIF &     0.556 & 0.744 & 0.650 & 0.651 & 0.594 & 0.515 & 0.597 & 0.589 & 0.610 \\
    FSIMc &   0.627 & 0.794 & 0.710 & 0.660 & 0.590 & 0.573 & 0.581 & 0.601 & 0.638 \\
    NLPD &    0.550 & 0.764 & 0.657 & 0.655 & 0.584 & 0.528 & 0.552 & 0.580 & 0.606 \\
    GMSD &    0.609 & 0.772 & 0.690 & 0.677 & 0.594 & 0.517 & 0.575 & 0.591 & 0.624 \\ \midrule
    DeepIQA & 0.703 & 0.794 & 0.748 & 0.660 & 0.582 & 0.585 & 0.598 & 0.606 & 0.654 \\
    PieAPP &  0.725 & 0.769 & 0.747 & 0.685 & 0.582 & 0.594 & 0.598 & 0.615 & 0.659 \\
    LPIPS &   0.760 & 0.828 & 0.794 & 0.705 & 0.605 & 0.625 & 0.630 & 0.641 & 0.692 \\
    DISTS &   \second{0.772} & 0.822 & \second{0.797} & \second{0.710} & 0.600 & 0.627 & 0.625 & 0.641 & 0.693 \\
    \midrule
    LPIPS+ & 0.756 & \second{0.833} & 0.795 & 0.706 & \second{0.606} & \second{0.630} & \second{0.631} & \second{0.643} & \second{0.694} \\
    \ours (ResNet50) & \best{0.805} & \best{0.843} & \best{0.824} & \best{0.724} & \best{0.616} & \best{0.662} & \best{0.634} & \best{0.659} & \best{0.714}
    \\
    \bottomrule
    \end{tabular}
    }
\end{table*}

\subsection{Comparison with NR Methods}

\begin{table}[t]
\centering
\caption{Quantitative comparison on \textbf{NR benchmarks}: CLIVE, KonIQ-10k and FLIVE.} \label{tab:nr_benchmark}
\resizebox{\linewidth}{!}{
\setlength{\tabcolsep}{1pt}
\begin{tabular}{lcccccc}
\toprule
& \multicolumn{2}{c}{CLIVE} & \multicolumn{2}{c}{KonIQ-10k}& \multicolumn{2}{c}{FLIVE} \\
\cmidrule(rl){2-3}
\cmidrule(rl){4-5}
\cmidrule(rl){6-7}
Methods & PLCC & SRCC &PLCC & SRCC &PLCC & SRCC 
\\ \midrule
DIIVINE~\cite{moorthy2011nssdiivine} & 0.591 & 0.588 & 0.558 & 0.546 & 0.186 & 0.092 \\
BRISQUE~\cite{2012brisque} & 0.629 & 0.629 & 0.685 & 0.681 & 0.341 & 0.303 \\ 
NIQE~\cite{2012niqe} & 0.493 & 0.451 & 0.389 & 0.377 & 0.211 & 0.288 \\ 
ILNIQE~\cite{zhang2015ilniqe} & 0.508 & 0.508 & 0.537 & 0.523 & 0.332 & 0.294 \\ 
PI~\cite{blau2018pi} & 0.521 & 0.462 & 0.488 & 0.457 & 0.334 & 0.170 \\ 
\midrule
PQR~\cite{zeng2018pqr} & 0.836 & 0.808 & - & - & - & - \\
MEON~\cite{ma2017meon} &  0.710 & 0.697 & 0.628 & 0.611 & 0.394 & 0.365\\
WaDIQaM~\cite{bosse2017wadiqam} & 0.671 & 0.682 & 0.807 & 0.804 & 0.467 & 0.455 \\
DBCNN~\cite{2020dbcnn} & 0.869 & \second{0.869} & 0.884 & 0.875 & 0.551 & 0.545 \\ 
HyperIQA~\cite{hyperiqa} & \second{0.882} & 0.859 & 0.917 & 0.906 & 0.602 & 0.544 \\ 
MetaIQA~\cite{zhu2020metaiqa} & 0.802 & 0.835 & 0.856 & 0.887 & 0.507 & 0.540 \\ 
TIQA~\cite{tiqa2021} & 0.861 & 0.845 & 0.903 & 0.892 & 0.581 & 0.541 \\
TReS~\cite{tres2022wacv} & 0.877 & 0.846 & \second{0.928} & 0.915 & 0.625 & 0.554 \\ 
MUSIQ~\cite{ke2021musiq} & - & - & \second{0.928} & \second{0.916} & \second{0.739} & \second{0.646} \\ \midrule
Ours (ResNet50) & \best{0.884} & \best{0.870} & \best{0.939} & \best{0.926} & 0.722 & 0.633 \\ 
std & $\pm$0.012  & $\pm$0.014 & $\pm$0.003 & $\pm$0.003 & - & - \\ 
\hdashline
\ours (Swin) & - & - & - & - & \best{0.745} & \best{0.652} \\
\bottomrule
\end{tabular}
}
\end{table}

\begin{table*}[t]
    \centering
    \caption{PLCC/SRCC scores of cross-dataset experiments with NR benchmarks.} \label{tab:cross_nr}
    \Large
    \resizebox{\linewidth}{!}{
    \renewcommand{\arraystretch}{1.2}
    \begin{tabular}{cccccccccc}
    \toprule
    Train on & \multicolumn{3}{c}{KonIQ-10k} & \multicolumn{3}{c}{FLIVE} & \multicolumn{3}{c}{SPAQ} \\ 
    \cmidrule(rl){2-4}
    \cmidrule(rl){5-7}
    \cmidrule(rl){8-10}
    Test on &  CLIVE & FLIVE & SPAQ & CLIVE & KonIQ-10k & SPAQ & CLIVE & KonIQ-10k & FLIVE \\ \midrule
    TReS & 0.8118/0.7771 & 0.513/0.4919 & 0.8624/0.8619 & 0.7213/0.7336 & 0.7507/0.7068 & 0.6137/0.7269 & -- & -- & -- \\ 
    MUSIQ & 0.8295/0.7889 & 0.5128/0.4978 & 0.8626/0.8676 & 0.8014/0.7672 & 0.7655/0.7084 & 0.8112/0.8436 & 0.8134/0.789 & 0.7528/0.6799 & 0.6039/0.5627 \\ \midrule
    \ours & \best{0.8389/0.8206} & \best{0.6272/0.5796} & \best{0.8791/0.8758} & \best{0.8140/0.7868} & \best{0.8008/0.7622} & \best{0.812/0.8479} & \best{0.8327/0.8128} & \best{0.8112/0.7632} & \best{0.6154/0.5653} \\ 
    \bottomrule
    \end{tabular}
    }
\end{table*}

\begin{table}[t] \centering
    \caption{Results on SPAQ dataset.} \label{tab:nr_spaq}
    \small
    \setlength{\tabcolsep}{18pt}
    \begin{tabular}{lcc}
    \toprule
    Method & PLCC & SRCC \\ \midrule
    DIIVINE~\cite{moorthy2011nssdiivine}  &  0.600 & 0.599 \\
    BRISQUE~\cite{2012brisque} & 0.817 & 0.809 \\ 
    ILNIQE~\cite{zhang2015ilniqe} & 0.721 & 0.713 \\ 
    PI~\cite{blau2018pi} & 0.724 & 0.709 \\ 
    \midrule
    Fang \etal \cite{fang2020spaq} & 0.909 & 0.908 \\
    DBCNN~\cite{2020dbcnn} & 0.915 & 0.911 \\ 
    MUSIQ~\cite{ke2021musiq} & \second{0.920} & \second{0.917} \\
    \midrule
    \ours (ResNet50) & \best{0.924} & \best{0.921} \\
    std & $\pm$0.002 & $\pm$0.003 \\
    \bottomrule
    \end{tabular}
\end{table}

\begin{table}[t]
    \caption{Results of KonIQ-10k using official split.}
    \label{tab:supp_koniq}
    \small
    \centering
    \setlength{\tabcolsep}{18pt}
    \begin{tabular}{lcc}
    \toprule
    Method & PLCC & SRCC \\ \midrule
    DIIVINE & 0.612 & 0.589 \\
    BRISQUE & 0.707 & 0.705  \\ \hdashline
    KonCept512~\cite{koniq10k} & \second{0.937} & 0.921 \\
    MUSIQ~\cite{ke2021musiq} & \second{0.937} & \second{0.924} \\
    \ours (ResNet50) & \best{0.941} & \best{0.928} \\
    \bottomrule
    \end{tabular}
\end{table}

\begin{table}[t] \centering
    \caption{Results on AVA dataset. ThemeAware$^\dagger$ uses extra theme labels.} \label{tab:nr_ava}
    \small
    \begin{tabular}{llcc}
    \toprule
    Method & Backbone & PLCC & SRCC \\ \midrule
    NIMA~\cite{talebi2018nima} & Inception-v2 & 0.636 & 0.612 \\
    PQR~\cite{zeng2018pqr} & ResNet101 & 0.720 & 0.719 \\ 
    Hosu \etal \cite{mlspf2019} & Inception-v2 & 0.757 & 0.756 \\
    ThemeAware$^\dagger$~\cite{jia2022themeaware} & Inception-v2 & \textcolor{gray}{0.775} & \textcolor{gray}{0.774} \\
    MUSIQ~\cite{ke2021musiq} & ViT-B/32 & 0.726 & 0.738 \\ 
    KD~\cite{hou2022kd} & ResNeXt101 & \second{0.770} & \second{0.770} \\ 
    \midrule
    \multirow{2}{*}{\ours} & ResNet50 & 0.733 & 0.733 \\
    & Swin & \best{0.790} & \best{0.791} \\ 
    \bottomrule
    \end{tabular}
\end{table}

NR-IQA is more challenging than FR-IQA due to the lack of references and the complexity of criteria. As discussed in related works, we split the NR datasets into two types: technical quality assessment and aesthetic quality assessment, as shown in \cref{tab:datasets}. We compare the proposed CFANet on both of these types in the following sections.

\subsubsection{Results on technical distortion benchmarks} There are mainly three NR datasets with authentic distortion, namely CLIVE (also known as the LIVE Challenge dataset), KonIQ-10k, and SPAQ, with the latter two being much larger than the first one. According to the results in \cref{tab:nr_benchmark} and \cref{tab:nr_spaq}, we can see that traditional approaches based on hand-crafted NSS features cannot handle natural images with complicated authentic distortions, while deep learning methods perform much better. In all three of these datasets, our model with a ResNet50 backbone outperforms existing CNN-based methods in both PLCC and SRCC. Our results are also better than MUSIQ, which is a purely vision transformer architecture. This indicates that the proposed CFANet is effective for authentic distortions even without reference images.  

\subsubsection{More results on KonIQ-10k}
Following previous works \cite{zhu2020metaiqa,ke2021musiq}, we report the results of 10 random splits on KonIQ-10k in \cref{tab:nr_benchmark}. However, \cite{koniq10k} provides a fixed split in their official codes\footnote{\url{https://github.com/subpic/koniq}}, and reports their results on it. We also report our results with the same setting in \cref{tab:supp_koniq}. We can observe that with a simple ResNet50 backbone, CFANet outperforms both KonCepth512 with inception-resnet-v2~\cite{szegedy2017inception} and MUSIQ with a vision transformer~\cite{vaswani2017attention}. This further proves the effectiveness and efficiency of the proposed CFANet.

\subsubsection{Results for aesthetic quality estimation} The AVA dataset is the primary benchmark for aesthetic evaluation. Since FLIVE has approximately 23\% overlap with images in the AVA dataset, we combine them for comparison. Unlike technical distortion, the assessment of image aesthetic quality pays more attention to the global feeling, where global semantics are more important than local textures. From the results in \cref{tab:nr_ava}, we can observe that ThemeAware significantly improves the results by introducing extra theme labels, and KD achieves better results by distilling semantic knowledge from multiple classification backbones. Since the proposed CFANet is mainly designed to better extract local distortions, its performance is expected to be worse than methods with more powerful classification backbones. However, CFANet with ResNet50 still achieves competitive results in both \cref{tab:nr_benchmark} and \cref{tab:nr_ava}, indicating that CFANet still preserves global semantic information well. We suspect that the residual connections in SA and CSA blocks enable CFANet to adaptively fuse global and local information. Next, we replace the ResNet50 backbone in CFANet with a relatively cheaper transformer backbone, namely the Swin transformer~\cite{liu2021swin}. From \cref{tab:nr_benchmark} and \cref{tab:nr_ava}, we can observe that CFANet-Swin outperforms the previous state-of-the-art methods on both FLIVE and AVA.  

\subsubsection{Cross dataset experiments.} We also conducted cross-dataset experiments on NR benchmarks to establish the robustness of our proposed method. 
\newline \indent \textbf{Experiment setting.} We used three NR datasets (KonIQ-10k, FLIVE, and SPAQ) from \cref{tab:datasets} for training. The CLIVE dataset is only used for testing, as it is relatively small, and the AVA dataset is an aesthetic dataset, thus not applicable in this context. Regarding KonIQ-10k and FLIVE, we utilized the official test split that contains approximately 2k and 7.3k images, respectively. Since SPAQ does not have an official split, we employed the entire dataset for testing, which contains approximately 11k images.
\newline \indent \textbf{Results.} As demonstrated in \cref{tab:cross_nr}, the proposed CFANet significantly outperforms other approaches. These results are consistent with the cross-dataset experiments on FR datasets in \cref{tab:fr_cross}, both of which highlight the advanced robustness and generalization capabilities of the proposed CFANet. 

\section{Ablation Study and Backbone Analysis}

In this section, we first present ablation experiments on the proposed components in CFANet, and then analyze the effects of different backbones on FR and NR tasks, respectively.

\subsubsection{Ablation of the proposed components} 
In \cref{tab:ablation_components}, we evaluate the proposed components in CFANet with a cross-dataset experiment, similar to \cref{tab:fr_cross}, as it does not require random splits and leads to a more fair comparison. The baseline model is a simple linear regression network with multi-scale features after global average pooling, and each proposed component is added sequentially. All model variants are trained on KADID-10k and tested on CSIQ and TID2013. We evaluate four components of CFANet: 1) Gated Local Pooling (GLP); 2) Self-Attention (SA); 3) Cross-scale Attention (CSA) and 4) Position embedding (Pos.). We can observe that all four components are beneficial to the results. Specifically, the GLP and SA blocks slightly improve the baseline performance. The CSA block brings the most significant improvement, which proves the effectiveness of top-down semantic propagation. The Pos. also contributes slightly to the final performance. The full CFANet makes significant improvements to the baseline. 

\begin{table*}[t]
    \centering
    \caption{Ablation study through cross dataset experiments for different components in CFANet. Experiments are done for both FR and NR datasets. (PLCC SRCC) scores are reported.}
    \label{tab:ablation_components}
    \setlength{\tabcolsep}{3pt}
    \begin{tabular}{c*{5}{c}cccccccc}
    \toprule
    \multirow{2}{*}{Model Index} & \multirow{2}{*}{ResNet50} & \multirow{2}{*}{GLP} & \multirow{2}{*}{SA} & \multirow{2}{*}{CSA} & \multirow{2}{*}{Pos.} & \multicolumn{4}{c}{KADID-10k (FR)} & \multicolumn{4}{c}{KonIQ-10k (NR)} \\
    \cmidrule(rl){7-10}
    \cmidrule(rl){11-14}
    & & & & & & \multicolumn{2}{c}{CSIQ} & \multicolumn{2}{c}{TID2013} & \multicolumn{2}{c}{CLIVE} & \multicolumn{2}{c}{SPAQ}\\ 
    \midrule
    \circled{1} & \cmark & & & & & 0.946 & 0.945 & 0.891 & 0.886 & 0.792 & 0.775 & 0.853 & 0.851 \\
    \circled{2} & \cmark & \cmark & & & & 0.952 & 0.952 & 0.894 & 0.885 & 0.808 & 0.801 & 0.861 & 0.860 \\
    \circled{3} & \cmark & \cmark & \cmark & & & 0.952 & 0.954 & 0.896 & 0.894 & 0.824 & 0.809 & 0.866 & 0.863 \\
    \circled{4} & \cmark & \cmark & \cmark & \cmark & & 0.963 & 0.965 & 0.912 & 0.908 & 0.836 & 0.817 & 0.874 & 0.872 \\
    \circled{5} & \cmark & \cmark & \cmark & \cmark & \cmark & 0.963 & 0.969 & 0.916 & 0.915 & 0.839 & 0.821 & 0.879 & 0.876 \\ \midrule
    \circled{a} & \cmark & Resize & \cmark & \cmark & \cmark & 0.961 & 0.961 & 0.913 & 0.910 & 0.834 & 0.814 & 0.868 & 0.865 \\
    \circled{b} & \cmark & \cmark & \cmark & Convolution fusion & \cmark & 0.958 & 0.960 & 0.910 & 0.908 & 0.830 & 0.813 & 0.865 & 0.862 \\
    \circled{c} & \cmark & \cmark & \cmark & Top layer guidance & \cmark & 0.956 & 0.957 & 0.905 & 0.903 & 0.821 & 0.806 & 0.864 & 0.860 \\ 
    \bottomrule
    \end{tabular}
\end{table*}

\subsubsection{Ablation with different variants} To further validate the effectiveness of our architecture design, we conduct experiments of the following three variants of CFANet:
\begin{itemize}
    \item \circled{a}~Replacing GLP with resize.
    \item \circled{b}~Replacing CSA with convolution fusion. 
    \item \circled{c}~Directly using top-layer feature to guide lowest-layer. 
\end{itemize}
According to the results presented in Table \ref{tab:ablation_components}, we can make the following observations about the overall performance: \circled{a} $>$ \circled{b} $>$ \circled{c}. From this, we can draw the following conclusions: 1) the proposed GLP is slightly superior to resize since GLP can more accurately and selectively capture local distortion information; 2) the proposed CSA outperforms convolution fusion, likely because the attention mechanism is more effective in aggregating features from the entire image; and 3) leveraging multi-scale semantic information is crucial for achieving optimal performance. These findings lend support to the effectiveness of the proposed modules.

\begin{figure}[t]
     \centering
     \includegraphics[width=\linewidth]{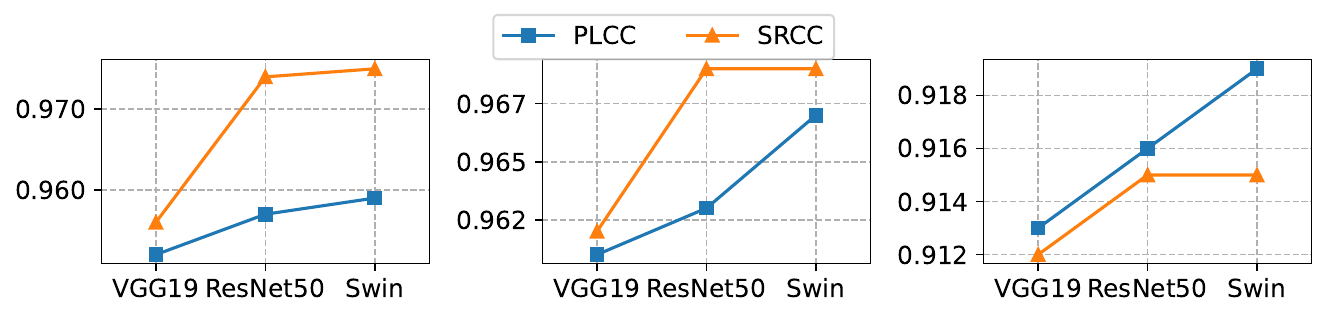} \\
     \makebox[0.31\linewidth]{\footnotesize (a) LIVE}\hfill
     \makebox[0.31\linewidth]{\footnotesize (b) CSIQ}\hfill
     \makebox[0.31\linewidth]{\footnotesize (c) TID2013}\hfill
     \caption{Results of different backbones on FR benchmarks.}
     \label{fig:ablation_backbone_fr}
     
     \centering
     \includegraphics[width=\linewidth]{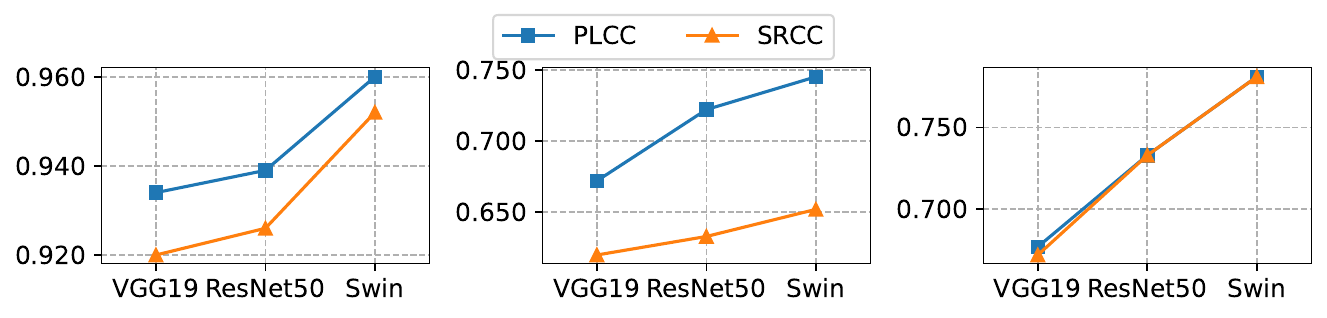} \\
     \makebox[0.31\linewidth]{\footnotesize (a) KonIQ-10k}\hfill
     \makebox[0.31\linewidth]{\footnotesize (b) FLIVE}\hfill
     \makebox[0.31\linewidth]{\footnotesize (c) AVA}\hfill
     \caption{Results of different backbones on NR benchmarks.}
     \label{fig:ablation_backbone_nr}
\end{figure}

\subsubsection{Performances with different backbones} In the previous experiments, we found that the backbone has a significant impact on the performance of aesthetic quality estimation. Therefore, we further evaluate how different backbones affect the performance on FR and NR benchmarks, respectively. We choose three representative backbones in our experiments, \ie, VGG19~\cite{iclr15vgg}, ResNet50, and Swin transformer, and the results are shown in \cref{fig:ablation_backbone_fr} and \cref{fig:ablation_backbone_nr}. We can observe that stronger backbones generally give better performance in both FR and NR benchmarks. However, the improvement between CFANet-Swin and CFANet-ResNet50 is much larger on NR benchmarks ($+0.02$) than on FR benchmarks ($+0.003$). We hypothesize that there are two main reasons: 1) the FR task relies more on the difference between distorted images and reference images, which is much easier to model, and simple ResNet50 is sufficient; 2) without reference images, the NR task needs to evaluate the global aesthetic quality, and transformers are good at learning global representation. Despite the differences, we are surprised to find that CFANet-VGG already outperforms most previous approaches on several FR and NR benchmarks. It proves the superiority of the proposed top-down framework to combine semantics with distortions in IQA.

\section{Conclusion}

In this work, we have proposed a top-down method, named as \emph{TOPIQ} for image quality assessment. Drawing inspiration from our understanding of the global-to-local processes of HVS, we hypothesize that semantic information is critical in guiding the perception of local distortions. By extending the widely used LPIPS method with feature re-weighting, we have discovered that current bottom-up techniques fail to exploit multi-scale features to their full potential as they neglect the importance of semantic guidance. To address this issue, we propose a heuristic top-down network, \ie, the coarse-to-fine attention network (CFANet), which effectively propagates multi-scale semantic information to low-level distortion features. The key element of CFANet is a novel cross-scale attention (CSA) mechanism that utilizes high-level features to guide the selection of semantically significant low-level features. We have also devised a gated local pooling (GLP) block to improve the efficiency of CSA. Lastly, we have conducted comprehensive experimental comparisons on various public benchmarks for both Full-Reference (FR) and No-Reference (NR) scenarios. Our proposed CFANet, with ResNet50 backbone, exhibits the best or highly competitive performance across all relevant benchmarks and is substantially more efficient than state-of-the-art approaches.


\bibliographystyle{IEEEtran}
\bibliography{iqa}

%

\end{document}